\documentclass[sigconf]{aamas} 

\usepackage{subcaption}
\usepackage{graphicx,wrapfig}
\usepackage{amsmath}
\usepackage{amsxtra}
\usepackage{amstext}
\usepackage{amsfonts}
\usepackage{amsthm}
\usepackage{dsfont}
\usepackage{mathtools}
\usepackage{adjustbox}

\usepackage{footnote}
\makesavenoteenv{tabular}

\usepackage{pifont}
\DeclareMathOperator{\KL}{\mathbb{K}\mathbb{L}}

\DeclarePairedDelimiterX{\infdivx}[2]{(}{)}{%
  #1\;\delimsize\|\;#2}
  \newcommand{\infdiv}{\infdivx}
\DeclareMathOperator{\EX}{\mathbb{E}}

\DeclareMathSymbol{\shortminus}{\mathbin}{AMSa}{"39}

\usepackage{balance} 

\renewcommand\footnotetextcopyrightpermission[1]{} 
\setcopyright{none}
\acmConference[AAMAS '22]{}
\copyrightyear{}
\acmYear{}
\acmDOI{}
\acmPrice{}
\acmISBN{}

\makeatletter
\def\@copyrightspace{\relax}
\makeatother

\acmSubmissionID{86}

\title[How to Sense the World]{\emph{How to Sense the World}:\\ Leveraging Hierarchy in Multimodal Perception for \\ Robust Reinforcement Learning Agents}

\author{Miguel Vasco}
\affiliation{%
  \institution{INESC-ID \& Instituto Superior T\'{e}cnico,\\ Universidade de Lisboa}
  \city{Lisbon}
  \country{Portugal}
}

\author{Hang Yin}
\affiliation{
  \institution{KTH Royal Institute of Technology}
  \city{Stockholm}
  \country{Sweden}}

\author{Francisco S. Melo}
\affiliation{%
  \institution{INESC-ID \& Instituto Superior T\'{e}cnico,\\ Universidade de Lisboa}
  \city{Lisbon}
  \country{Portugal}
}

\author{Ana Paiva}
\affiliation{%
  \institution{INESC-ID \& Instituto Superior T\'{e}cnico,\\ Universidade de Lisboa}
  \city{Lisbon}
  \country{Portugal}
}

\renewcommand{\shortauthors}{Vasco, et al.}

\begin{abstract}
This work addresses the problem of \emph{sensing the world}: how to learn a multimodal representation of a reinforcement learning agent's environment that allows the execution of tasks under incomplete perceptual conditions. To address such problem, we argue for \emph{hierarchy} in the design of representation models and contribute with a novel multimodal representation model, MUSE. The proposed model learns a hierarchy of representations: low-level modality-specific representations, encoded from raw observation data, and a high-level multimodal representation, encoding joint-modality information to allow robust state estimation. We employ MUSE as the perceptual model of deep reinforcement learning agents provided with multimodal observations in Atari games. We perform a comparative study over different designs of reinforcement learning agents, showing that MUSE allows agents to perform tasks under incomplete perceptual experience with minimal performance loss. Finally, we also evaluate the generative performance of MUSE in literature-standard multimodal scenarios with higher number and more complex modalities, showing that it outperforms state-of-the-art multimodal variational autoencoders in single and cross-modality generation.
\end{abstract}


\newcommand{\BibTeX}{\rm B\kern-.05em{\sc i\kern-.025em b}\kern-.08em\TeX}

\begin{document}


\pagestyle{fancy}
\fancyhead{}


\maketitle 


\section{Introduction}
\label{Section:intro}

\begin{figure}[t]
\centering
\includegraphics[width=0.9\columnwidth]{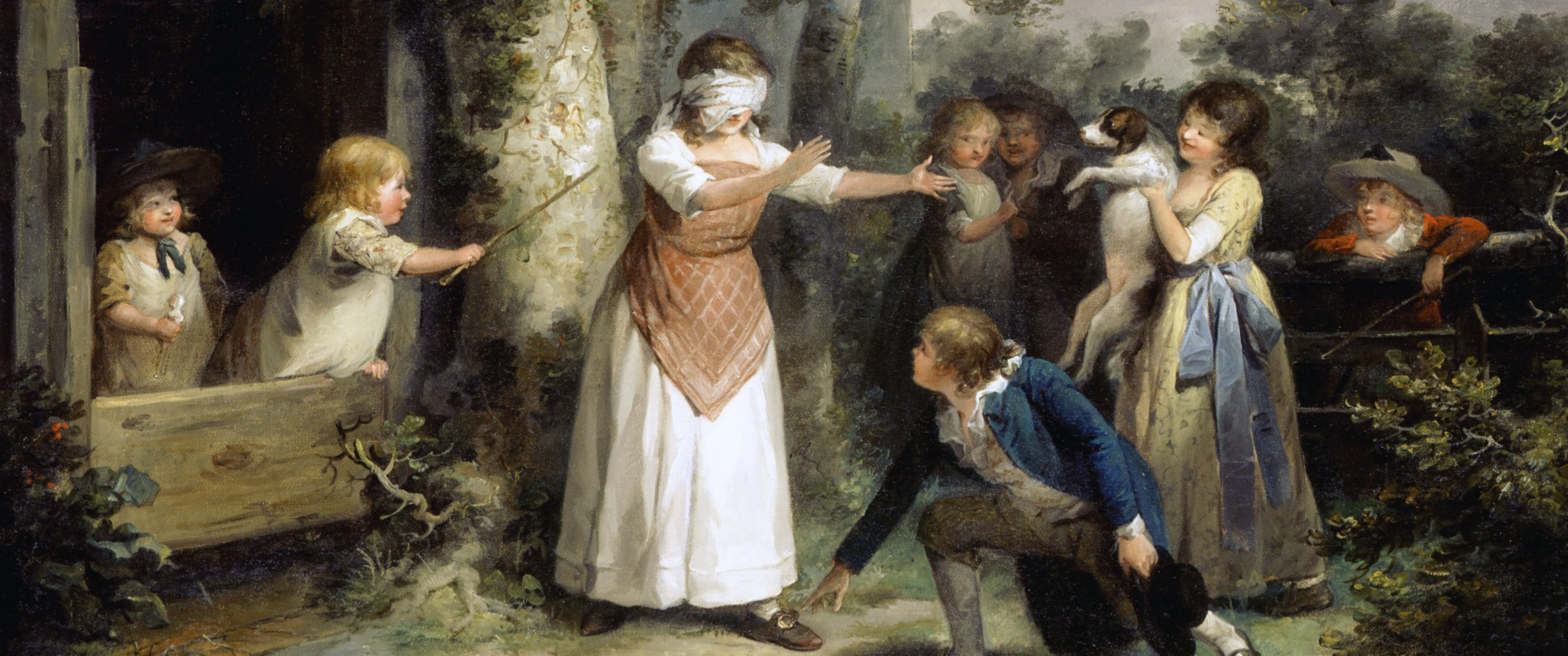}
\caption{Acting under incomplete perceptual conditions. Adapted from George Morland's \emph{Blind Man's Buff} (1788).}
\label{fig:intro}
\vspace{-1ex}
\end{figure} 

Perceiving the world is a core skill for any autonomous agent. To act efficiently, the agent must be able to collect, process and interpret the perceptual information provided by its environment. This information can be of considerable complexity, such as the high-dimensional pixel information provided by the agent's camera or the sounds collected by the agent's microphones. To reason efficiently over such information, agents are often endowed with mechanisms to encode low-dimensional internal representations of their complex perceptions. Such representations enable more efficient learning of how to act~\citep{gelada2019deepmdp,zhang2019solar} and facilitate the adaptation to distinct (albeit similar) domains~\citep{higgins2017darla,ha2018world}. 

\begin{figure*}[t]
    \centering
    \begin{subfigure}[b]{0.31\textwidth}
        \centering
        \includegraphics[height=3cm]{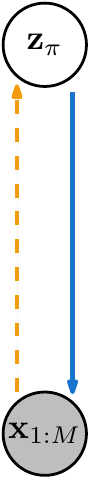}
        \caption{VAE.}
        \label{fig:muse:vae}
    \end{subfigure}%
    \hfill 
    \begin{subfigure}[b]{0.31\textwidth}
        \centering
         \includegraphics[height=3cm]{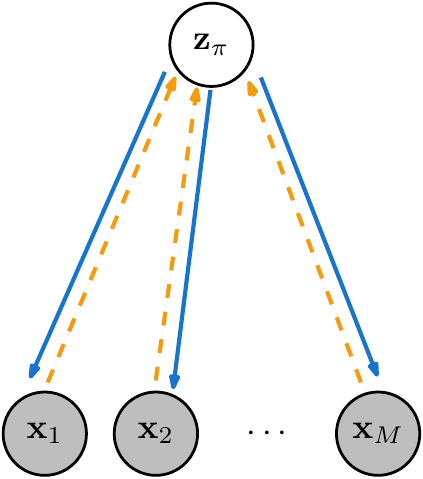}
        \caption{Multimodal VAE.}
        \label{fig:muse:mvae}
    \end{subfigure}
    \hfill 
     \begin{subfigure}[b]{0.31\textwidth}
        \centering
        \includegraphics[height=3cm]{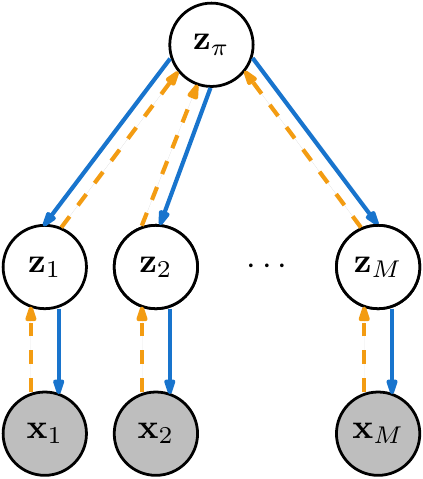}
       \caption{MUSE.}
        \label{fig:muse:model}
    \end{subfigure}
    \caption{Explicit representation models for RL agents in multimodal scenarios: (a) VAE model with a flat latent representation $\mathbf{z}_\pi$ and fused observations $\mathbf{x_{[1:M]}}$~\cite{higgins2017darla,ha2018world,hafner2019dream}; (b) Multimodal VAE (MVAE) with a flat latent representation $\mathbf{z}_\pi$~\cite{silva2020playing}; (c) the proposed MUSE model, with hierarchical levels: modality-specific ($\mathbf{z}_{1:M}$) and a multimodal ($\mathbf{z}_\pi$) representations. We highlight the encoders (orange, dashed) and decoders (blue, full) networks of the models.}
    \label{fig:muse}
    \vspace{-2ex}
\end{figure*}

In this work we pivot on the design of representations for sensory information. In particular, we address the problem of \emph{sensing the world}: how can we process an agent's sensory information is such a way that it is robust to changes in the perceptual conditions of the environment (such as the removal of input modalities), mimicking animal behavior~\citep{partan2017multimodal} and the human experience (Fig.~\ref{fig:intro}).

Recently, deep reinforcement learning (RL) agents have shown remarkable performance in complex control tasks when provided with high-dimensional observations, such as in Atari Games~\citep{arulkumaran2017brief, mnih2015human,lillicrap2015continuous}. Early approaches considered training a controller, instantiated as a neural network, directly over sensory information (e.g. pixel data), thus learning an \emph{implicit} representation of the observations of the agent within the structure of the controller. However, in a multimodal setting, such agents struggle to act when provided with incomplete observations: the missing input information flowing through the network degrades the quality of the agent's output policy. To provide some robustness to changing sensory information, other agents learn \emph{explicit} representation models of their sensory information, often employing variational autoencoder (VAE) models, before learning how to perform a task in their environment. These explicit representations allow agents to perceive their environment and act in similar environments, such as in the DARLA framework~\cite{higgins2017darla}, or to act from observations generated by the representation itself, such as in the World Models framework~\cite{ha2018world}.

In this work, we focus on the design of explicit representation models that are robust to changes in the perceptual conditions of the environment (i.e. with unavailable modalities at execution time) for RL agents in multimodal scenarios. In such conditions, traditional approaches employing VAE models with a fixed \emph{fusion} of the agent's sensory information struggle to encode a suitable representation: the missing information flowing within the network degrades the quality of the encoded state representation and, subsequently, of the agent's policy~\cite{collier2020vaes}. To address such problem, multimodal variational autoencoders (MVAE) have been recently employed to learn representations in a multimodal setting~\citep{silva2020playing}. By considering the \emph{independent} processing of each modality, these models are able to overcome the problems of fusion solutions, as no missing information is propagated through the network~\citep{silva2020playing}. However, by design, the models encode information from all modalities into a single \emph{flat} representation space of finite capacity, often leading the agent to neglect information from lower-dimensional modalities~\citep{shi2019variational}, hindering the estimation of the state of the environment when high-dimensional observations are unavailable.

To provide robust state estimation to the agent, regardless of the nature of the provided sensory observations, we argue for designing perceptual models with \emph{hierarchical} representation levels: raw observation data is processed and encoded in low-level, modality-specific representations, which are subsequently merged in a high-level, multimodal representation. By accounting for the intrinsic complexity of each input modality at a low-level of the model, we are able to encode a multimodal state representation that better accounts for all modalities and is robust to missing sensory information. We instantiate such design in a novel framework, contributing the \emph{Multimodal Unsupervised Sensing} (MUSE) model.

We evaluate MUSE as an explicit representation model for RL agents in a recently-proposed deep reinforcement learning scenario, where an agent is provided with multimodal observations of Atari games~\cite{silva2020playing}. We perform a comparative study against other architectures of reinforcement learning agents and show that only RL agents employing MUSE are able to perform tasks with missing observations at test time, incurring in a minimal performance loss. Finally, we assess the performance of MUSE in scenarios with increasing complexity and number of modalities and show, quantitatively and qualitatively, that it outperforms state-of-the-art MVAEs. 

In summary, the main contributions of this work are threefold: \emph{(i)} MUSE, a novel representation model for multimodal sensory information of RL agents that considers hierarchical representation levels; \emph{(ii)} a comparative study on different architectures of RL agents showing that, with MUSE, RL agents are able to perform tasks with missing sensory information, with minimal performance loss; \emph{(iii)} an evaluation of MUSE in literature-standard multimodal scenarios of increasing complexity and number of modalities showing that it outperforms state-of-the-art multimodal variational autoencoders.

\section{Background}
\label{Section:background}

Reinforcement learning (RL) is a computational framework for decision-making through trial-and-error interaction between an agent and its environment \cite{sutton1998reinforcement}. RL problems can be formalized using Markov decision processes (MDP) that describe sequential decision problems under uncertainty. An MDP can be instantiated as a tuple $\mathcal{M} = (\mathcal{X},\mathcal{A},P,r,\gamma)$, where $\mathcal{X}$ and $\mathcal{A}$ are the (known) state and action spaces,
respectively. When the agent takes an
action $a\in\mathcal{A}$ while in state $x\in\mathcal{X}$, the world transitions to
state $y\in\mathcal{X}$ with probability $P(y\mid x,a)$ and the agent receives
an immediate reward $r(x,a)$.  Finally, the discount factor
$\gamma \in [0, 1)$ defines the relative importance of present and future rewards for the agent.

\begin{figure*}[t]
    \centering
    \begin{subfigure}[b]{0.31\textwidth}
        \centering
         \includegraphics[height=3.1cm]{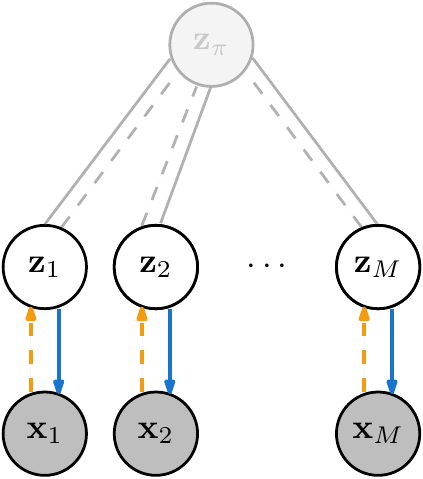}
        \caption{Modality-specific factors $\mathbf{z}_{1:M}$.}
        \label{fig:muse:bottom}
    \end{subfigure}
    \hfill 
    \begin{subfigure}[b]{0.31\textwidth}
        \centering
        \includegraphics[height=3.1cm]{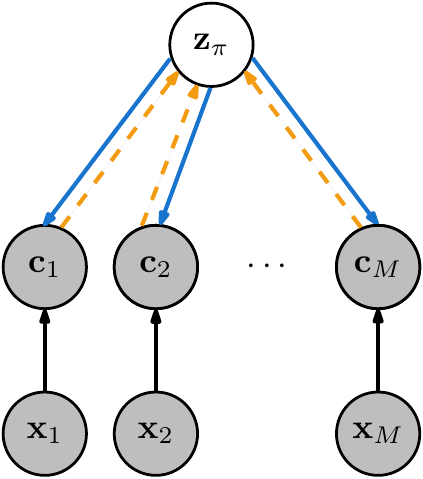}
        \caption{Multimodal factors $\mathbf{z}_{\pi}$.}
        \label{fig:muse:top}
    \end{subfigure}%
    \hfill 
     \begin{subfigure}[b]{0.31\textwidth}
        \centering
        \includegraphics[height=3.1cm]{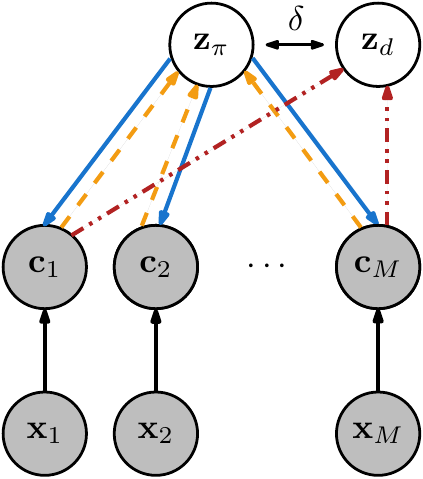}
       \caption{ALMA training scheme.}
        \label{fig:muse:alma}
    \end{subfigure}
    \caption{Learning hierarchical representations in the MUSE framework: (a) the modality-specific factors $\mathbf{z}_{1:M}$, employing the loss in Eq.~\eqref{Eq:bottom_loss}; (b) the multimodal factors $\mathbf{z}_{\pi}$, encoded from the low-dimensional representations $\mathbf{c}_{1:M}$, employing the loss in Eq.~\eqref{Eq:top_loss} (c) the Average Latent Multimodal Approximation (ALMA) training scheme, where we approximate the complete multimodal latent variable $\mathbf{z}_{\pi}$ with partial multimodal latent variables $\mathbf{z}_{d}$, encoded from observations with missing modalities.}
    \label{fig:muse:learning}
    \vspace{-2ex}
\end{figure*}

The reward function $r$, typically unknown to the agent, encodes the goal of the agent in its environment, whose dynamics are described by $P$, often unknown to the agent as well. Through a trial-and-error approach, the agent aims at learning a policy $\pi : \mathcal{X} \times \mathcal{A} \to \left[ 0, 1 \right]$, where $\pi(a \mid x)$ is the probability of performing action $a\in \mathcal{A}$ in state $x \in \mathcal{X}$, that maximizes the expected reward collected by the agent. Such a policy can be found from the optimal $Q$-function, defined for every state-action pair $\left( x, a \right) \in \mathcal{X} \times \mathcal{A}$ as
\begin{equation}
  Q^*(x,a) = r(x, a) + \gamma \sum_{y\in\mathcal{X}} P(y \mid x, a) \max_{a'\in\mathcal{A}} Q^*(y, a').
\end{equation}
To compute the optimal $Q$-function, multiple methods can be employed~\cite{sutton1998reinforcement}, such as
$Q$-learning~\cite{watkins89phd}. Recently, RL has been applied to scenarios where the agent collects high-dimensional observations from its environment, leading to new methods and extensions of classical reinforcement learning methods~\cite{arulkumaran2017brief}. For discrete action spaces the Deep $Q$ Network (DQN) was proposed, a variant of the $Q$-learning algorithm, that employs a deep neural network to approximate the optimal $Q$-function~\cite{mnih2015human}. For continuous action spaces the Deep
Deterministic Policy Gradient (DDPG) algorithm allows agents to perform complex control tasks~\cite{lillicrap2015continuous}, for example in robotic manipulation~\cite{vecerik2017leveraging}. However, in a multimodal setting, both algorithms learn an implicit representation of the fused sensory information provided to the agent, which may hinder their performance in scenarios with missing modality information.

\section{MUSE}
\label{Section:Methods}

In this work, we consider that a reinforcement learning agent is provided  with information regarding the state of environment through $M$ different channels, \mbox{$\{\mathbf{x}_1, \ldots, \mathbf{x}_{M}\}$}, where $\mathbf{x}_m$ is the information provided by an input ``modality''. Each modality may correspond to a different type of information (e.g., image, sound).

To design RL agents for such scenario, we can employ several distinct architectures: one can employ a neural-network controller that learns a policy $\pi$ directly from the fixed fusion of all observations $\mathbf{x}_{[1:M]}$, thus learning an implicit representation of sensory information, as in DQN or DDPG~\citep{mnih2015human,lillicrap2015continuous}. One can also employ the same fusion mechanism to learn an explicit representation of the agent's observations using a VAE model (Fig.~\ref{fig:muse:vae}) and, subsequently, learn a policy over the flat latent representation $\mathbf{z}_{\pi}$~\citep{higgins2017darla,ha2018world,hafner2019dream}. Another architecture uses a multimodal VAE (Fig.~\ref{fig:muse:mvae}), able to consider the independent processing of each modality to encode $\mathbf{z}_{\pi}$~\citep{yin2017associate,suzuki2016joint,wu2018multimodal,shi2019variational}. However, two issues arise from such design choices:
\begin{itemize}
    \item The fusion solution employed by both implicit and explicit representation models is not robust to missing observations: the missing input propagates through the network reducing its performance, thus leading to a inaccurate state representation and to an incorrect policy;
    \item  The flat representation solution with independent processing encodes information from all modalities into a finite-capacity representation, often disregarding the information provided by low-dimensional observations, as shown in~\citet{shi2019variational}. Thus, this option struggles to encode a robust state representation solely from low-dimensional modalities.
\end{itemize} 
To provide robust state representation, regardless of the number and nature of the available modalities, we argue for considering a \emph{hierarchical} architecture that accommodates:
\begin{itemize}
    \item Low-level {\em modality-specific} representations, encoding information specific to each modality in (low-level) latent variables whose capacity (dimensionality) is individually defined considering the intrinsic complexity of each modality;
    
    \item A high-level \emph{multimodal} representation, merging information from the available low-level representations. By using low-dimensional representations as input, the multimodal representation can better balance the information provided by the distinct modalities and, thus, encode a robust representation regardless of the nature of the available observations.
\end{itemize}

\subsection{Learning Hierarchical Representations} 

We instantiate the previous hierarchical design in the \emph{Multimodal Unsupervised Sensing} (MUSE) model, depicted in Fig.\ref{fig:muse:model}. To train the modality-specific representations $\mathbf{z}_{1:M}$, we assume that each modality $\mathbf{x}_m$ is generated by a corresponding latent variable $\mathbf{z}_m$ (Fig.~\ref{fig:muse:bottom}). We can employ a loss function $\ell_{b}(\mathbf{x}_{1:M})$, similar to the single-modality VAE loss~\cite{kingma2013auto}, to learn a set of independent, single-modality, generative models $p_{\theta}(\mathbf{x}_{1:M}) = \prod_{m=1}^{M} p_{\theta}(\mathbf{x}_{m})$, i.e.,
\begin{equation}
\begin{split}
     \ell_{b}(\mathbf{x}_{1:M}) = \sum_{m=1}^{M} & \bigg(\EX_{q_{\phi}(\mathbf{z}_m | \mathbf{x}_m)} \left[ \lambda_m \log p_{\theta}(\mathbf{x}_m | \mathbf{z}_m) \right] \\
     -& \alpha_m \KL \infdiv{q_{\phi}(\mathbf{z}_m|\mathbf{x}_m)}{p(\mathbf{z}_m)}\bigg),
\end{split}
   \label{Eq:bottom_loss}
\end{equation}
where $\lambda_m$ affects the reconstruction of modality-specific data and $\alpha_m$ controls the regularization of the corresponding latent variable.

At a high-level, we encode modality-specific information to learn a multimodal representation, $\mathbf{z}_{\pi}$. As shown in Fig.~\ref{fig:muse:top}, we employ the modality-specific encoder networks to sample an efficient low-level representation of modality data $\mathbf{c}_{1:M} \sim \left[q_{\phi}(\mathbf{z}_{1:M}|\mathbf{x}_{1:M})\right]$ \footnote{At test time, we compute deterministically the low-level representation $\mathbf{c}_{1:M} = \EX \left[ q_{\phi}(\mathbf{z}_{1:M}|\mathbf{x}_{1:M})\right]$}. We assume that the representations $\mathbf{c}_{1:M}$ are generated by a multimodal latent variable $\mathbf{z}_{\pi}$ and employ a loss function $\ell_{t}(\mathbf{x}_{1:M})$, similar to the standard MVAE loss~\cite{wu2018multimodal}, to learn the multimodal representation, 

\begin{equation}
\begin{split}
   \ell_{t}(\mathbf{x}_{1:M}) =  \EX_{q_{\phi}(\mathbf{c}_{1:M}| \mathbf{x}_{1:M})} & \bigg( - \beta \KL \infdiv{q_{\phi}(\mathbf{z}_{\pi} \mid \mathbf{c}_{1:M})}{p(\mathbf{z}_{\pi})}\\
   + \sum_{m=1}^{M} & \EX_{q_{\phi}(\mathbf{z}_{\pi} \mid \mathbf{c}_{1:M})} \left[\gamma_m\log p_{\theta} (\mathbf{c}_{m}\mid \mathbf{z}_{\pi})\right]\bigg),
   \end{split}
      \label{Eq:top_loss}
\end{equation}
where $\gamma_m$ affects the reconstruction of the modality-specific representations and $\beta$ controls the regularization of the multimodal latent variable.

\begin{figure*}[t]
    \centering
    \begin{subfigure}[b]{0.49\textwidth}
        \centering
        \includegraphics[height=1.0cm]{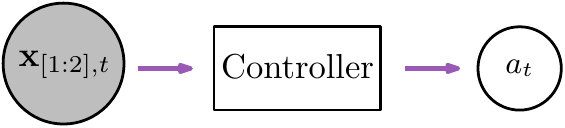}
        \caption{\textbf{Multimodal Controller}}
        \label{fig:atari:multimodal_RL}
    \end{subfigure}%
    \hfill 
    \begin{subfigure}[b]{0.49\textwidth}
        \centering
        \includegraphics[height=1.0cm]{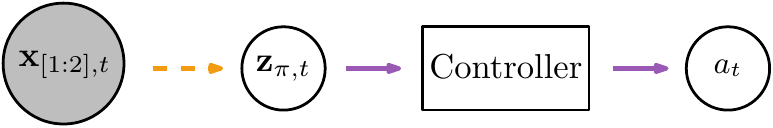}
        \caption{VAE + Controller}
        \label{fig:atari:vae_rl}
    \end{subfigure}
    
    \vspace{3ex}
    
     \begin{subfigure}[b]{0.49\textwidth}
        \centering
        \includegraphics[height=1.6cm]{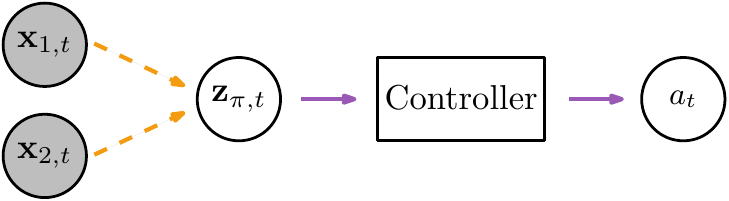}
        \caption{\textbf{MVAE + Controller}}
        \label{fig:atari:mvae_rl}
    \end{subfigure}%
    \hfill
    \begin{subfigure}[b]{0.49\textwidth}
        \centering
        \includegraphics[height=1.6cm]{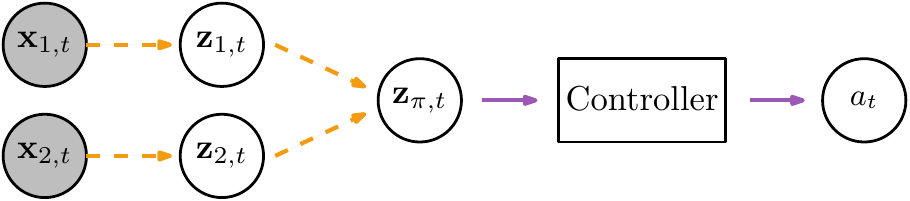}
        \caption{MUSE + Controller}
        \label{fig:atari:muse_rl}
    \end{subfigure}
    \caption{Designing reinforcement learning agents provided with  multimodal observations $\mathbf{x}_{1:M}$: (a) the Multimodal Controller solution; (b) the VAE + Controller solution; (c) the MVAE + Controller solution; (d) our proposed MUSE + Controller solution. We highlight the representation learning (orange, dashed) and the policy learning (purple, full) components of the agents.}
    \label{fig:atari:models}
\end{figure*}

\subsection{Multimodal Training Scheme}

To encode a multimodal representation that is both agnostic to the nature of the modalities and scalable to large number of modalities, we approximate the joint-modality posterior distribution ${p(\mathbf{z}_{\pi} \mid \mathbf{c}_{1:M})}$ using a product-of-experts (PoE), where,
\begin{equation}
    p(\mathbf{z}_{\pi} \mid \mathbf{c}_{1:M}) = p(\mathbf{z}_{\pi}) \prod_{m=1}^{M} q(\mathbf{z}_{\pi} \mid \mathbf{c}_{m}).
\end{equation}
This solution is able to scale to a large number of modalities~\cite{wu2018multimodal}: assuming that both the prior and posteriors are Gaussian distributions, the product-of-experts distribution is itself a Gaussian distribution with mean ${\mu = \left(\sum_m \mu_m T_m\right)\left(\sum_m T_m\right)^{-1}}$ and covariance ${\Sigma = \left(\sum_m T_m\right)^{-1}}$, where $T_m = \Sigma_m^{-1}$ and $\Sigma_m$ is the covariance of $q(\mathbf{z}_{\pi} \mid \mathbf{c}_{m})$. However, the original PoE solution is prone to learning overconfident experts, neglecting information from low-dimensional modalities, as shown in~\cite{shi2019variational}.

To address this issue, we introduce the \emph{Average Latent Multimodal Approximation} (ALMA) training scheme for PoE. With ALMA, we explicitly enforce the similarity between the multimodal distribution encoded from all modalities with the ``partial'' distributions encoded from observations with missing modalities, as depicted in Fig.~\ref{fig:muse:alma}. In particular, during training we encode a latent distribution $q(\mathbf{z}_{\pi} \mid \mathbf{c}_{1:M})$ that considers all modalities and, in addition, we also encode distributions with missing modality information, $q(\mathbf{z}_{d} \mid \mathbf{x}_{d})$, one for every possible combination of modalities. For example, in scenarios with two input modalities ($\mathbf{x}_1, \mathbf{x}_2$) we encode ${D=2}$ partial distributions, corresponding to ${\mathbf{z}_{d}^{1} \sim  q(\cdot \mid \mathbf{c}_1)}$ and ${\mathbf{z}_{d}^{2} \sim  q(\cdot \mid \mathbf{c}_2)}$. To encode a multimodal representation robust to missing modalities, we force all the multimodal distributions to be similar by including $|D|$ additional loss terms, yielding a final loss function
\begin{equation}
\begin{split}
\ell(\mathbf{x}_{1:M}) =& \,  \ell_{b}(\mathbf{x}_{1:M}) + \ell_{t}(\mathbf{x}_{1:M}) \\
+ &\frac{\delta}{|D|}\sum_{d \in D}\KL^{\star} \infdiv{q_{\phi}(\mathbf{z}_{\pi} \mid \mathbf{c}_{1:M})}{q_{\phi}(\mathbf{z}_{d} \mid \mathbf{c}_{d})},
\end{split}
\label{Eq:MUSE-loss}
\end{equation}
where the parameter $\delta$ governs the impact of the approximation loss term and $\KL^{\star}$ is the symmetrical KL-divergence, following~\cite{yin2017associate}. Employing the loss function of \eqref{Eq:MUSE-loss}, we train both representation levels concurrently in the same data pass through the model.%
\footnote{We stop the gradients of the top loss \eqref{Eq:top_loss} from propagating to the bottom-level computation graph by cloning and detaching the codes $\mathbf{c}_{1:M}$ sampled from the modality-specific distributions.} 
In Appendix we show how the ALMA term plays a fundamental role in addressing the overconfident expert phenomena of PoE solutions. 

\subsection{Learning To Act}

To employ MUSE as an sensory representation model for RL agents, we follow the three-step approach of~\citep{silva2020playing}: initially, we train MUSE on a previously-collected dataset of joint-modality observations  $\mathcal{D}(\mathbf{x}_{1:M})=\left\lbrace \mathbf{x}_{1:M}^0, \ldots, \mathbf{x}_{1:M}^N\right\rbrace$, using the loss function of Eq.~\eqref{Eq:MUSE-loss}. After training MUSE, we encode the perceptual observations of the RL agent in the multimodal latent state $\mathbf{z}_{\pi}$ to learn a policy $\pi : \mathcal{Z} \to \mathcal{A}$ that maps the latent states $\mathbf{z}_{\pi} \in \mathcal{Z}$ to actions of the agent $a \in \mathcal{A}$. To learn such a policy $\pi$ over the latent state, one can employ any continuous-state space reinforcement learning algorithm, such as DQN~\cite{mnih2015human} or DDPG~\cite{lillicrap2015continuous}.

\begin{figure*}[t]
    \centering
    \begin{subfigure}[b]{0.5\textwidth}
        \centering
        \includegraphics[height=4.7cm]{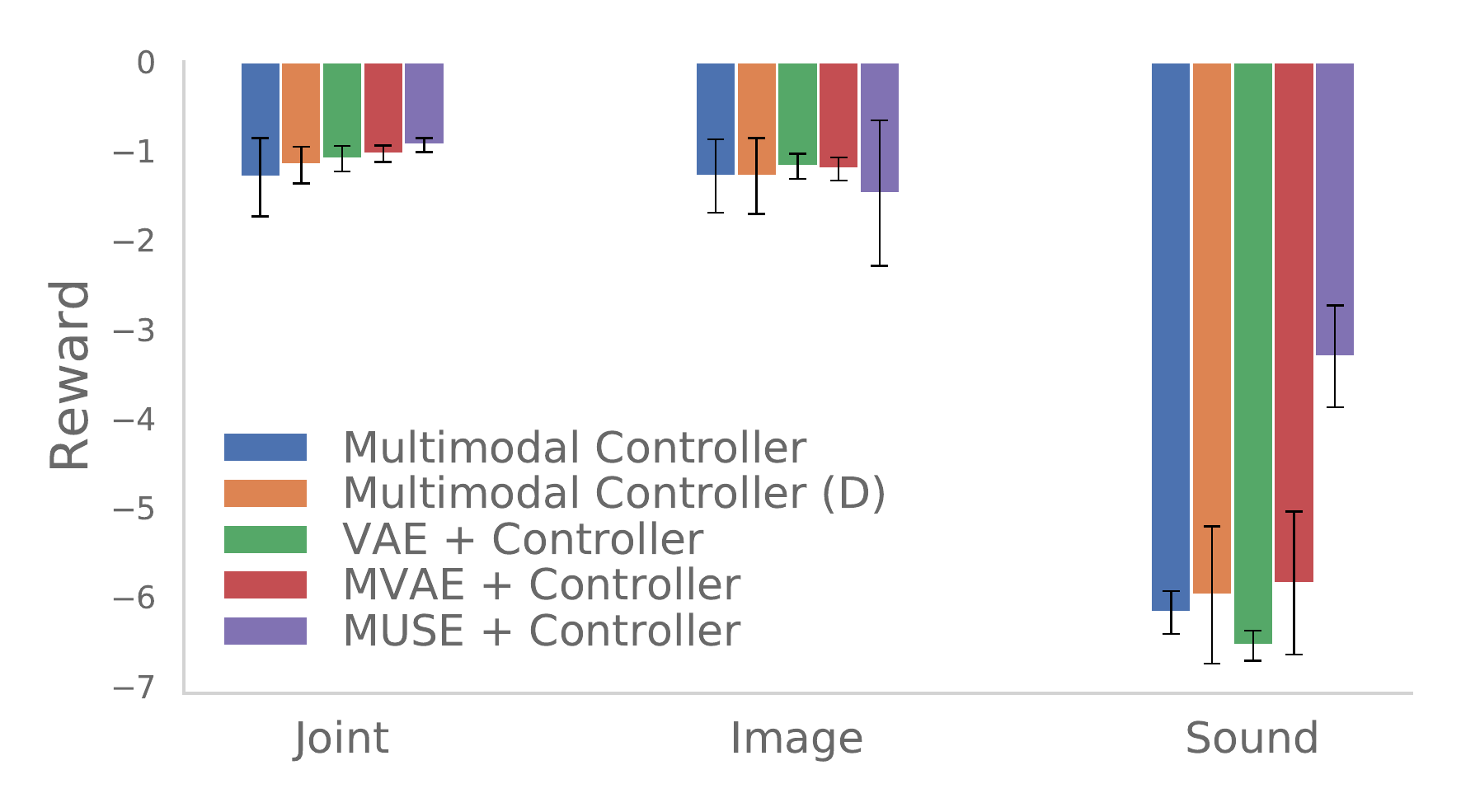}
        \caption{Pendulum}
        \label{fig:atari:pendulum_results}
    \end{subfigure}%
    \hfill
     \begin{subfigure}[b]{0.5\textwidth}
        \centering
        \includegraphics[height=4.7cm]{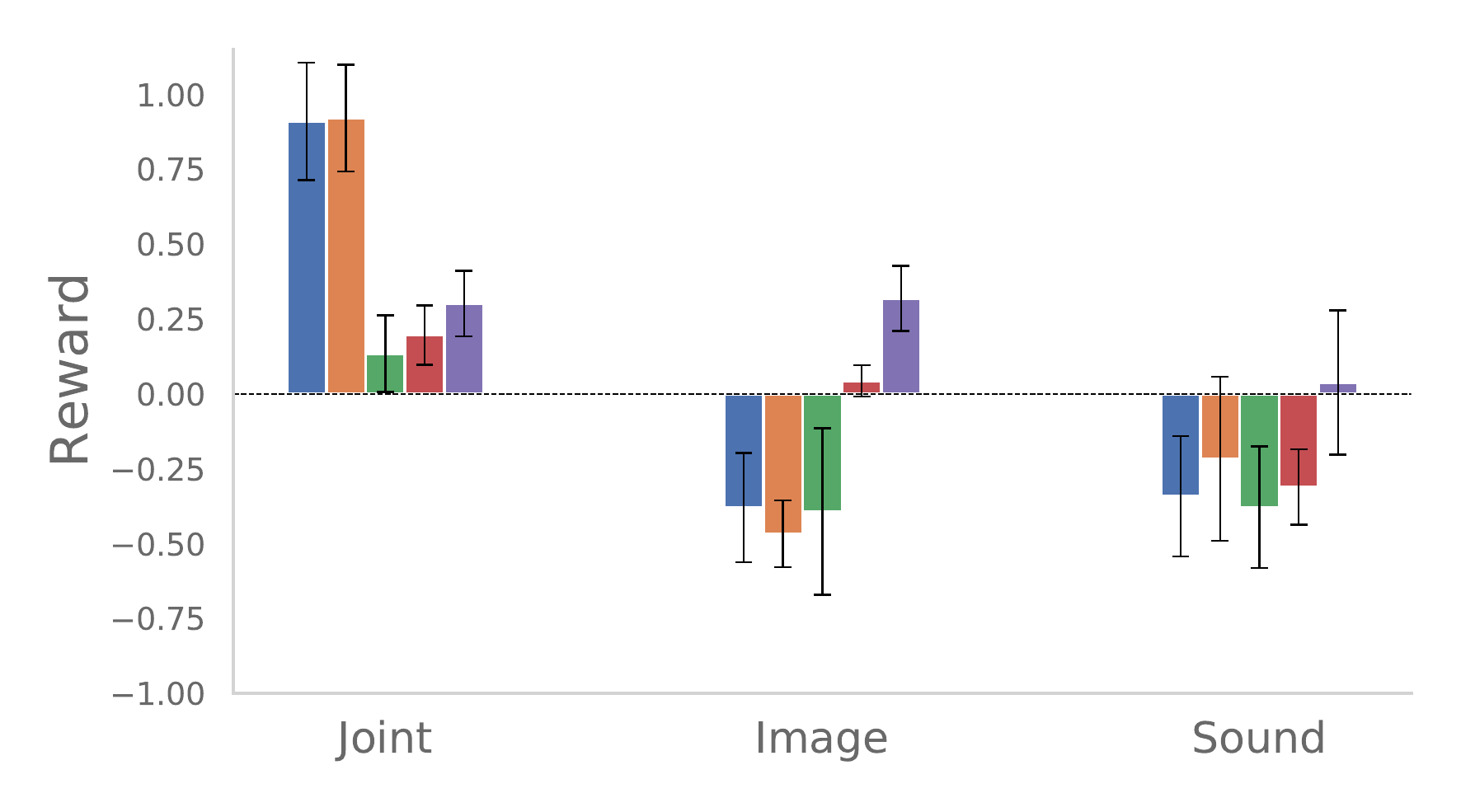}
        \caption{\textsc{Hyperhot}}
        \label{fig:atari:hyperhot_results}
    \end{subfigure}%
    \caption{Performance after zero-shot policy transfer in (a) the Pendulum scenario; and (b) the \textsc{Hyperhot} scenario. At test time, the agent is provided with either image ($\mathbf{x}_\text{I}$), sound ($\mathbf{x}_\text{S}$), or joint-modality ($\mathbf{x}_\text{I}, \mathbf{x}_\text{S}$) observations (Higher is better). All results averaged over 10 randomly seeded runs. Complete numerical results are presented in Appendix.}
    \label{fig:atari:comparison}
\end{figure*}

\section{Evaluation}
\label{Section:Evaluation}

We evaluate MUSE addressing the following two questions: \textbf{(i)} What is the performance of an RL agent that employs MUSE as a sensory representation model when provided with observations with missing modality information? \textbf{(ii)}  What is the standalone performance of MUSE as a generative model in scenarios with larger number and more complex modalities?

To address \textbf{(i)} we perform on Section~\ref{Section:eval:rl} a comparative study of different architectures of RL agents against a novel agent that employs MUSE as a sensory representation model. We evaluate the agents in the recently proposed multimodal Atari game scenario, where an agent is provided with multimodal observations of the game~\cite{silva2020playing}. The results show that---with MUSE--- the RL agents are able to perform tasks with missing modality information at test time, with minimal performance loss. To gain further insight into these results, we also address \textbf{(ii)}, evaluating the standalone performance of MUSE against other state-of-the-art representation models across different literature-standard multimodal scenarios. We provide quantitative and qualitative results on Section~\ref{Section:eval:generative} that show that MUSE outperforms other models in single and cross-modality generation. Code available at \url{https://github.com/miguelsvasco/muse}.

\subsection{Acting Under Incomplete Perceptions}
\label{Section:eval:rl}

We evaluate the performance of an RL agent that employs MUSE as a sensory representation model in a comparative study of different literature-standard agent architectures. In particular, we evaluate how the nature of the state representation (implicit vs explicit), the method of processing sensory information (fusion vs independent) and the type of representation model employed (flat vs hierarchical) allows agents to act under incomplete perceptions. As shown in Fig.~\ref{fig:atari:models}, we consider different design choices for RL agents:
\begin{itemize}
    \item \emph{Multimodal Controller} (Fig~\ref{fig:atari:multimodal_RL}), where we train a controller directly over the multimodal observations, in an end-to-end fashion (implicit, fusion, flat). This is the case of models such as DQN~\cite{mnih2015human} and DDPG~\cite{lillicrap2015continuous};
    \item \emph{VAE + Controller} (Fig~\ref{fig:atari:vae_rl}), where we initially train a representation model of the (fused) multimodal observations of the environment using a VAE and afterwards train the controller over the state representation $\mathbf{z}_{\pi}$ (explicit, fusion, flat). This is the case of frameworks such as World Models~\cite{ha2018world}, DARLA~\cite{higgins2017darla} and Dreamer~\cite{hafner2019dream};
    \item \emph{MVAE + Controller} (Fig~\ref{fig:atari:mvae_rl}), where we initially train a representation model of sensory information using the MVAE model~\cite{wu2018multimodal} and afterwards train the controller over the state representation $\mathbf{z}_{\pi}$ (explicit, independent, flat). This is the case of the models in Silva et al.~\cite{silva2020playing};
    \item Our proposed \textbf{MUSE + Controller} (Fig~\ref{fig:atari:muse_rl}), where we employ MUSE to learn a representation model of the environment and subsequently train a controller over the state representation $\mathbf{z}_{\pi}$ (explicit, independent, hierarchical);
\end{itemize}
To evaluate the role of the training scheme in the agent's performance, we also include a variation of the implicit agent \emph{Multimodal Controller (D)} that employs a training method similar to MUSE, in which modalities are randomly dropped while learning the policy.

We evaluate the agents on two recently-proposed multimodal scenarios for deep reinforcement learning%
\footnote{The multimodal Atari games are taken from \url{https://github.com/miguelsvasco/multimodal-atari-games}}: the multimodal Pendulum~\cite{brockman2016openai} and the \textsc{HyperHot} scenarios, described in Appendix. We adopt the same RL algorithms, training hyper-parameters and network architectures of~\cite{silva2020playing}, due to the similarity in evaluation scenarios. We compare the performance of the RL agents when directly using the policy learned from joint-modality observations in scenarios with possible missing modalities, without any additional training. Fig~\ref{fig:atari:pendulum_results} and \ref{fig:atari:hyperhot_results} summarize the total reward collected per episode, for the Pendulum and {\sc Hyperhot} scenarios, respectively. Results are averaged over 100 episodes and 10 randomly seeded runs. Numerical results shown in Appendix.

\begin{figure*}[t]
    \centering
    \begin{subfigure}[b]{0.45\textwidth}
        \centering
        \includegraphics[height=4.8cm]{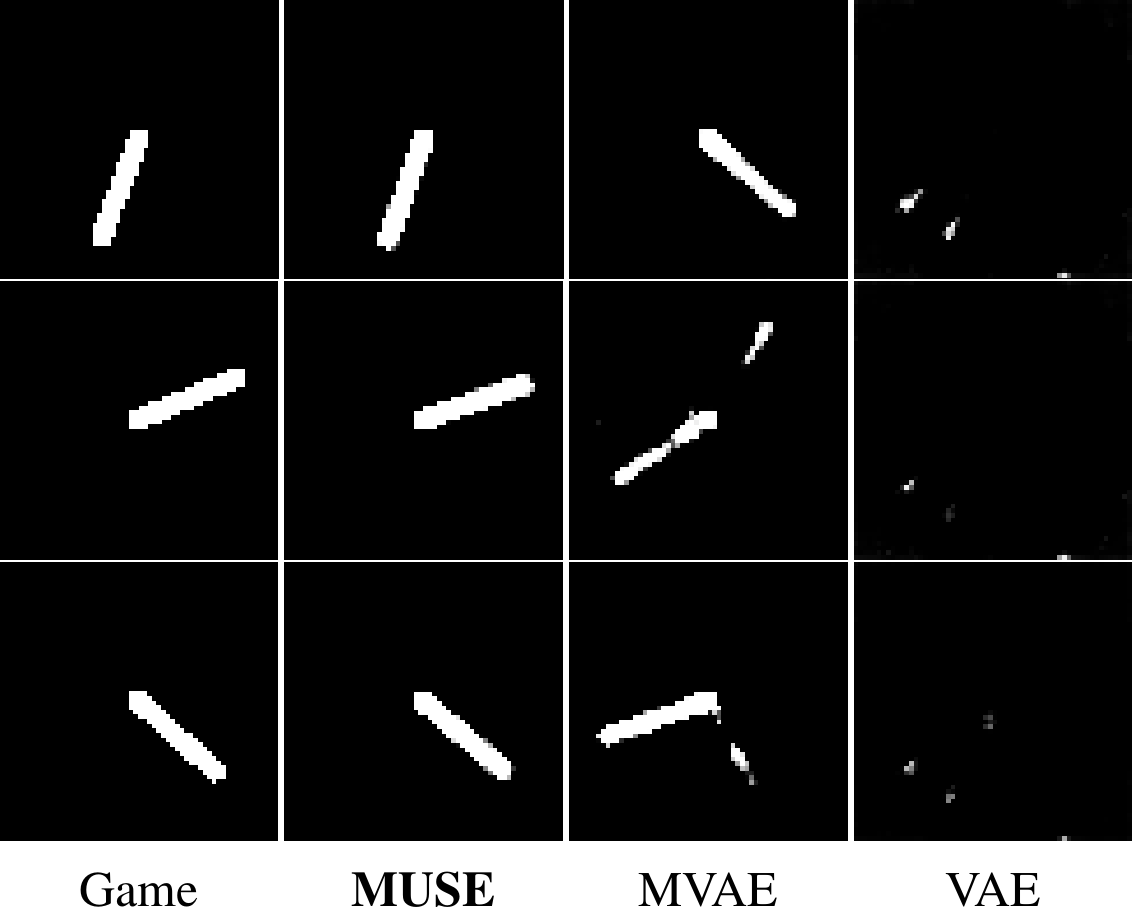}
        \caption{Pendulum}
        \label{fig:rl:eval:pendulum}
    \end{subfigure}%
    \hfill 
    \begin{subfigure}[b]{0.45\textwidth}
        \centering
        \includegraphics[height=4.8cm]{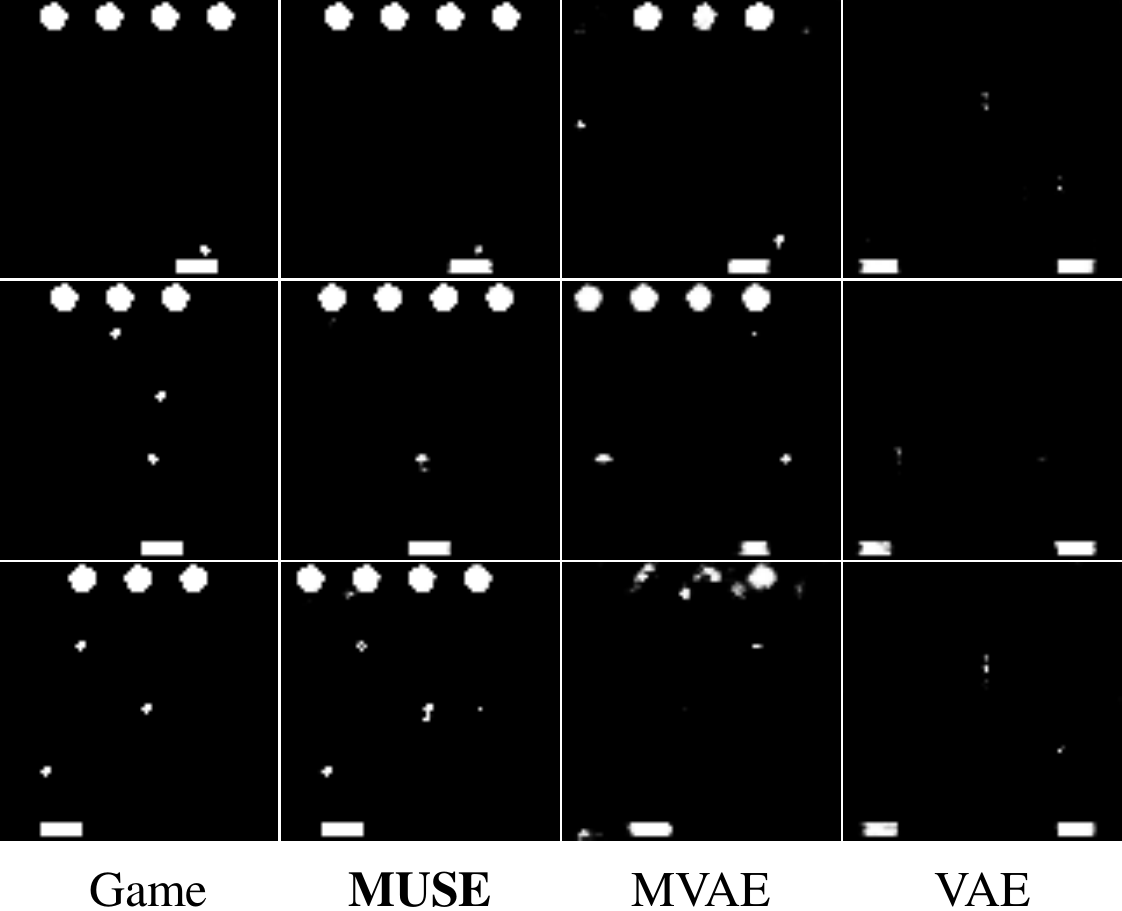}
        \caption{\textsc{Hyperhot}}
        \label{fig:rl:eval:hyperhot}
    \end{subfigure}%
    \caption{Cross-modality image generation from sound information in the multimodal Atari Game scenarios. MUSE is the only \emph{explict} representation model able to generate coherent image information from sounds. Best viewed with zoom.}
    \label{fig:rl:eval}
\end{figure*}

The results show that the MUSE + Controller agent is the only agent able to act robustly in incomplete perceptual conditions, regardless of the modalities available. In the Pendulum scenario, the performance of our agent is similar when provided with image or sound observations, as both modalities hold information to fully describe the state of the environment. In the {\sc Hyperhot} scenario, the sound modality is unable to describe the complete state of the environment and, as such, the MUSE + Controller agent provided with image observations outperforms the one provided with sound observations. However, even in the latter case our agent is able to successfully complete the majority of episodes, shown by the positive average reward.

\subsubsection{\emph{Implicit} vs. \emph{Explicit}}

The results in Fig.~\ref{fig:atari:comparison} regarding the performance of the agents when provided with joint-modality observations highlight the benefits (and challenges) of learning an explicit state representation. In the \textsc{Hyperhot} scenario, the implicit representation agent Multimodal Controller, which learns a policy directly from data, outperforms all other agents when provided with joint-modality observations. This result reveals the challenge of learning an accurate state representation in complex scenarios, suitable for RL tasks. However, in the simpler Pendulum scenario, the explicit representation agents are able to learn a robust low-dimensional state representation, performing on par with the Multimodal Controller agent when provided with joint-modality information.

\subsubsection{\emph{Fusion} vs. \emph{Independent}}

The Multimodal Controller and the VAE + Controller agents struggle to act when provided with only sound observations in both scenarios. Such behaviour confirms the robustness issue raised in Section~\ref{Section:Methods}: solutions that consider the fixed fusion of perceptions, both with implicit and explicit representations, are unable to robustly estimate the state of the environment when the agent is provided with incomplete perceptions. Moreover, the results of the variations with dropout show that the improved performance of the independent models is not due to their training scheme: the improvement by considering dropout with fused observations is limited for the Multimodal Controller (D) agent. 

\subsubsection{\emph{Flat} vs. \emph{Hierarchical}}

The MVAE + Controller agent also struggles to act when provided only with sound, despite an improvement in performance in comparison with the fusion solutions. This highlights the issue raised in Section~\ref{Section:Methods} regarding multimodal representation models with a flat latent variable: by employing a single representation space (of finite capacity) to encode information from all modalities, the model learns overconfident experts, neglecting information from lower-dimensional modalities~\citep{shi2019variational}. On the other hand, the hierarchical design of MUSE allows to balance the complexity of each modality by encoding a joint-representation from low-dimensional, modality-specific, representations.

\subsubsection{Qualitative Evaluation}

We can evaluate qualitatively the performance of \emph{explicit} representation models under incomplete perceptual experience by observing the image state generated from sound information, as shown in Fig.~\ref{fig:rl:eval}. The samples show that the MUSE + Controller solution is the only able to encode a robust estimation of state from low-dimensional information. In the Pendulum scenario, the MUSE agent is able to perfectly reconstruct the image observation of the game state, when only provided with sound information. This is coherent with the results in Fig.~\ref{fig:atari:comparison} regarding the similar performance when the agent is provided observations from either modality. In the \textsc{Hyperhot} scenario, sound information is unable to perfectly account for the image state of the environment: the images generated from sound information are unable to precisely describe the state of the environment, e.g. the number and position of enemies. This is also coherent with the results of Fig.~\ref{fig:atari:comparison} regarding the lower performance by the agent when provided with sound observations in comparison to the agent provided with image observations. However, the image reconstructions show that, even when encoded from sound, the multimodal representation contains fundamental information for the actuation of the agent, such as the position of projectiles near the agent. On the other hand, in both scenarios, the VAE + Controller and the MVAE + Controller agents are unable to reconstruct the (image) game state from sound information, evidence of their lack of robustness to missing modalities.

\begin{table*}[t]
\centering
\caption{Standard metrics for generative performance in the different datasets (best results in bold). We used 5000 importance samples for the MNIST dataset and 1000 importance samples for the CelebA dataset and the MNIST-SVHN scenario. All results averaged over 5 independent runs. Higher is better.}
\begin{subtable}{0.77\textwidth}
\centering
\caption{MNIST ($\mathbf{x}_1$ - Image; $\mathbf{x}_2$ - Label) }
\begin{adjustbox}{width=\columnwidth,center}
\begin{tabular}{@{}lccccc@{}}
\toprule
 Model & $\,\,\,\log p(\mathbf{x}_1)$ & $\,\,\,\log p(\mathbf{x}_2)$ & $\,\,\,\log p(\mathbf{x}_1,  \mathbf{x}_2)$ & $\,\,\,\log p(\mathbf{x}_1 \mid \mathbf{x}_2)$ & $\,\,\,\log p(\mathbf{x}_2 \mid \mathbf{x}_1)$ \\ \midrule
\textbf{MUSE (Ours)} & $\bf -24.06 \pm 0.03$  & $\bf -2.41 \pm 0.02$  & $-34.72 \pm 1.47$  & $-33.97 \pm 1.16$   & $\bf -\phantom{1}4.73 \pm 0.15$  \\
MVAE & $-24.45 \pm 0.14$  & $\bf -2.38 \pm 0.01$  & $\bf -25.46 \pm 0.28$  & $\bf -29.23 \pm 1.00$  & $-\phantom{1}9.60 \pm 0.38$  \\
MMVAE & $-25.96 \pm 0.11$ & $-2.82 \pm 0.07$ & - & $-39.94 \pm 0.48$  & $ -10.01 \pm 0.23$  \\  \bottomrule
\end{tabular}
\end{adjustbox}
\end{subtable}

\vspace{2ex}

\begin{subtable}{0.77\textwidth}
\centering
\caption{CelebA ($\mathbf{x}_1$ - Image; $\mathbf{x}_2$ - Attributes)}
\begin{adjustbox}{width=\columnwidth,center}
\begin{tabular}{@{}lccccc@{}}
\toprule
 Model & $\,\,\,\log p(\mathbf{x}_1)$ & $\,\,\,\log p(\mathbf{x}_2)$ & $\,\,\,\log p(\mathbf{x}_1,  \mathbf{x}_2)$ & $\,\,\,\log p(\mathbf{x}_1 \mid \mathbf{x}_2)$ & $\,\,\,\log p(\mathbf{x}_2 \mid \mathbf{x}_1)$ \\ \midrule
\textbf{MUSE (Ours)} & $\bf -163.63 \pm \phantom{1}0.13$   & $-19.69 \pm 0.09$   & $-288.25 \pm 3.70$  & $\bf -394.65 \pm 2.22$ & $\bf -\phantom{1}47.48 \pm 0.46$   \\
MVAE & $-165.88 \pm \phantom{1}0.40$  & $\bf -15.21 \pm 0.04$  & $\bf -181.56 \pm 0.44$  & $-471.50 \pm 1.67$  & $-\phantom{1}72.36 \pm 0.53$  \\
MMVAE & $-548.11 \pm 10.46$  & $-28.43 \pm 0.25$  & - & $-787.24 \pm 4.07$   & $-108.50 \pm 0.76$   \\  \bottomrule
\end{tabular}
\end{adjustbox}
\end{subtable}

\vspace{2ex}

\begin{subtable}{0.77\textwidth}
\centering
\caption{MNIST-SVHN ($\mathbf{x}_1$ - MNIST image; $\mathbf{x}_2$ - SVHN image; $\mathbf{x}_3$ - Label)}
\begin{adjustbox}{width=\columnwidth,center}
\begin{tabular}{@{}lccccc@{}}
\toprule
 Model & $\,\,\,\log p(\mathbf{x}_1)$ & $\,\,\,\log p(\mathbf{x}_2)$ & $\,\,\,\log p(\mathbf{x}_1| \mathbf{x}_2)$ & $\,\,\,\log p(\mathbf{x}_1 \mid \mathbf{x}_2, \mathbf{x}_3)$ & $\,\,\,\log p(\mathbf{x}_2 \mid \mathbf{x}_1)$ \\ \midrule
\textbf{MUSE (Ours)} & $\bf -24.23 \pm 0.02$   & $\bf -36.03 \pm 0.04$   & $\bf -39.93 \pm 1.45$  & $\bf -37.66 \pm 1.42$ & $-\phantom{1}64.99 \pm 3.35$   \\
MVAE & $-24.34 \pm 0.07$  & $\bf -36.17 \pm 0.22$  & $-43.44 \pm 0.19$  & $-40.96 \pm 0.86$  & $\bf -\phantom{1}55.51 \pm 0.18$  \\
MMVAE & $-28.21 \pm 0.06$  & $-41.33 \pm 0.23$  & $-54.84 \pm 0.47$ & - & $-166.14 \pm 0.76$   \\  \bottomrule
\end{tabular}
\end{adjustbox}
\end{subtable}
\label{Table:standard_metrics}
\end{table*}

\subsection{Generative Performance of MUSE}
\label{Section:eval:generative}

We now evaluate the potential of MUSE in learning a multimodal representation in scenarios with more complex and numerous modalities available to the agent. We do so by evaluating the generative performance of our model against two state-of-the-art multimodal variational autoencoders,%
\footnote{We selected these models as both are agnostic to the nature of the modalities and scalable to large number of modalities, similarly to MUSE (as shown in Section~\ref{Section:RW})}
the MVAE~\cite{wu2018multimodal} and MMVAE~\cite{shi2019variational},%
in literature-standard multimodal scenarios: the MNIST dataset~\cite{lecun1998gradient}, the CelebA~\cite{liu2015faceattributes} and the MNIST-SVHN scenario~\cite{netzer2011reading}.

We employ the authors' publicly available code\footnote{The MVAE model is taken from \url{https://github.com/mhw32/multimodal-vae-public} and the MMVAE model is taken from \url{https://github.com/iffsid/mmvae}.} 
and training loss functions, without importance-weighted sampling, as well as the suggested hyper-parameters (when available). In each scenario, we employ the same representation capacity (total dimensionality of representation spaces) in all models.  Code, model architectures and training hyper-parameters are presented in Appendix.

We compute standard likelihood-based metrics (marginal and joint), using both the joint variational posterior and the single variational posterior, averaged over 5 independently-seeded runs. Our results are summarized in Table~\ref{Table:standard_metrics} and we present image samples generated from label information in Fig.~\ref{fig:standard_cmi_samples} (more in Appendix). 

\subsubsection{Single-Modality Performance}
In the two-modality scenarios, MUSE outperforms all baselines in terms of the image marginal likelihood, $\log p(\mathbf{x}_1)$. Regarding the attribute marginal likelihood $\log  p(\mathbf{x}_2)$, MUSE performs on par with MMVAE in the MNIST dataset, despite employing a modality-specific latent representation 16 times smaller. The results in the three-modality scenario again show that MUSE outperforms the baselines in single-modality generation. The hierarchical design of MUSE allows the adjustment of the capacity of each modality-specific latent space to the inherent complexity of the corresponding modality.

\subsubsection{Joint-Modality Performance}
Regarding the joint-modality performance, ${\log p(\mathbf{x}_1, \mathbf{x}_2)}$, in MUSE, $\mathbf{z}_{\pi}$ encodes low-dimensional codes abstracted from modality data. This leads to a multimodal reconstruction process that loses some information for higher-dimensional modalities (e.g., image), yet accentuates distinctive features that define the observed phenomena (e.g., digit class), leading to lower generative performance. This phenomena can also be seen in reconstruction from higher latent variables in single-modality hierarchical generative models~\cite{havtorn2021hierarchical}. However, as seen in Section~\ref{Section:eval:rl}, such abstraction leads to improved state representation when the agent is provided with complex observations.

\subsubsection{Cross-Modality Performance}
\begin{figure*}[t]
    \centering
    \begin{subfigure}[b]{0.24\textwidth}
        \centering
        \includegraphics[height=4cm]{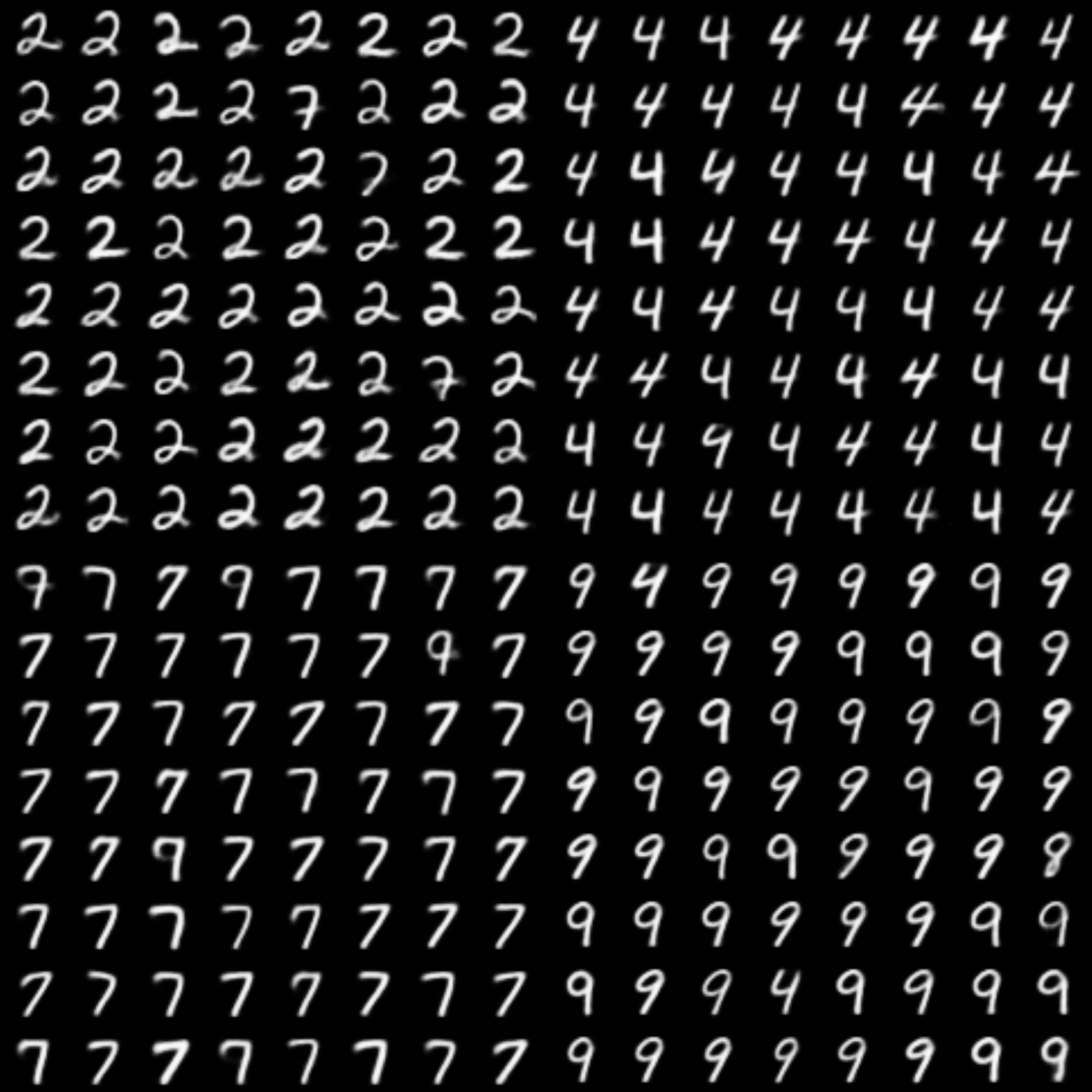}
        \caption{\textbf{MUSE (Ours)}}
        \label{fig:mnist_MUSE}
    \end{subfigure}%
    \hfill 
    \begin{subfigure}[b]{0.24\textwidth}
        \centering
        \includegraphics[height=4cm]{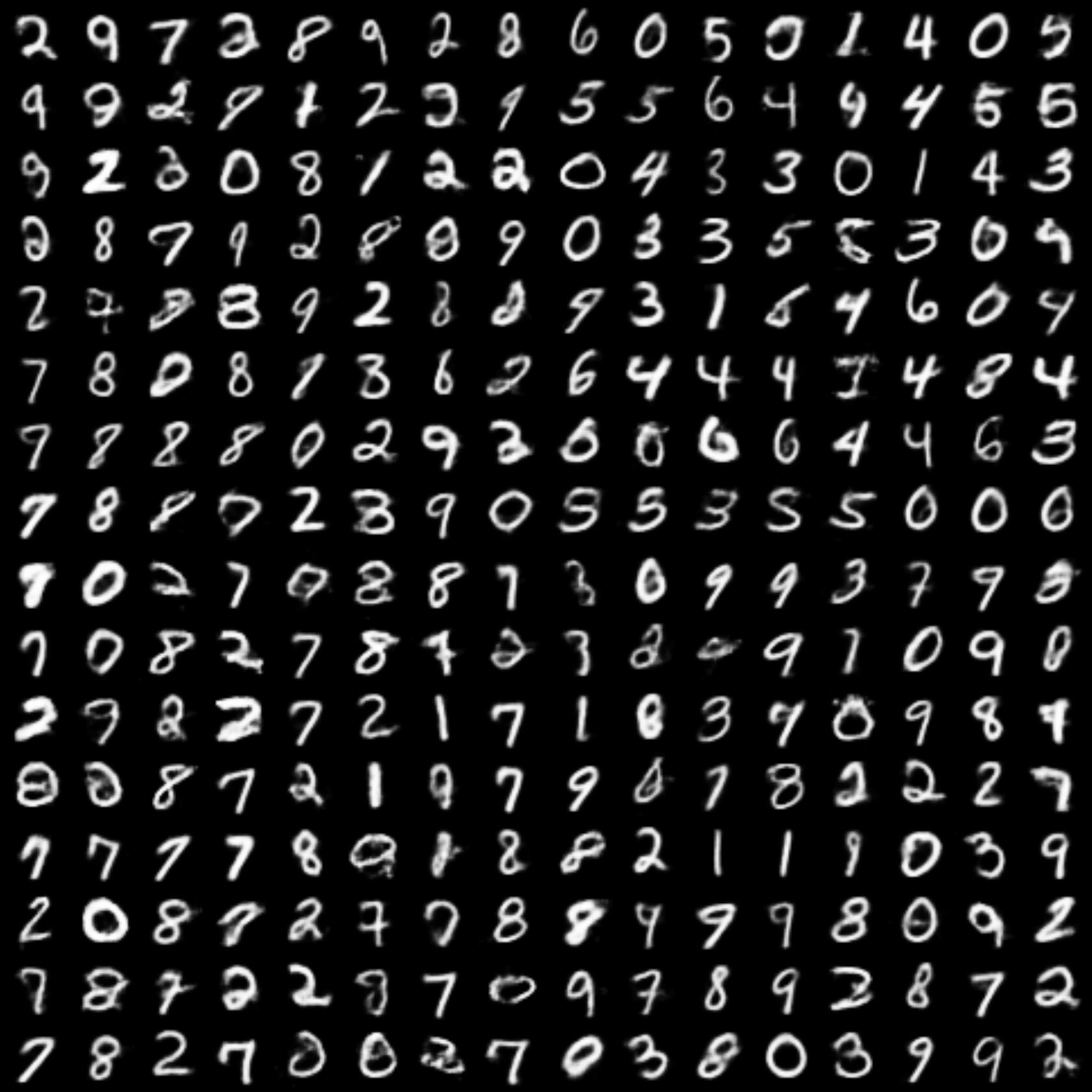}
        \caption{MVAE}
        \label{fig:mnist_mvae}
    \end{subfigure}
    \hfill
     \begin{subfigure}[b]{0.24\textwidth}
        \centering
        \includegraphics[height=4cm]{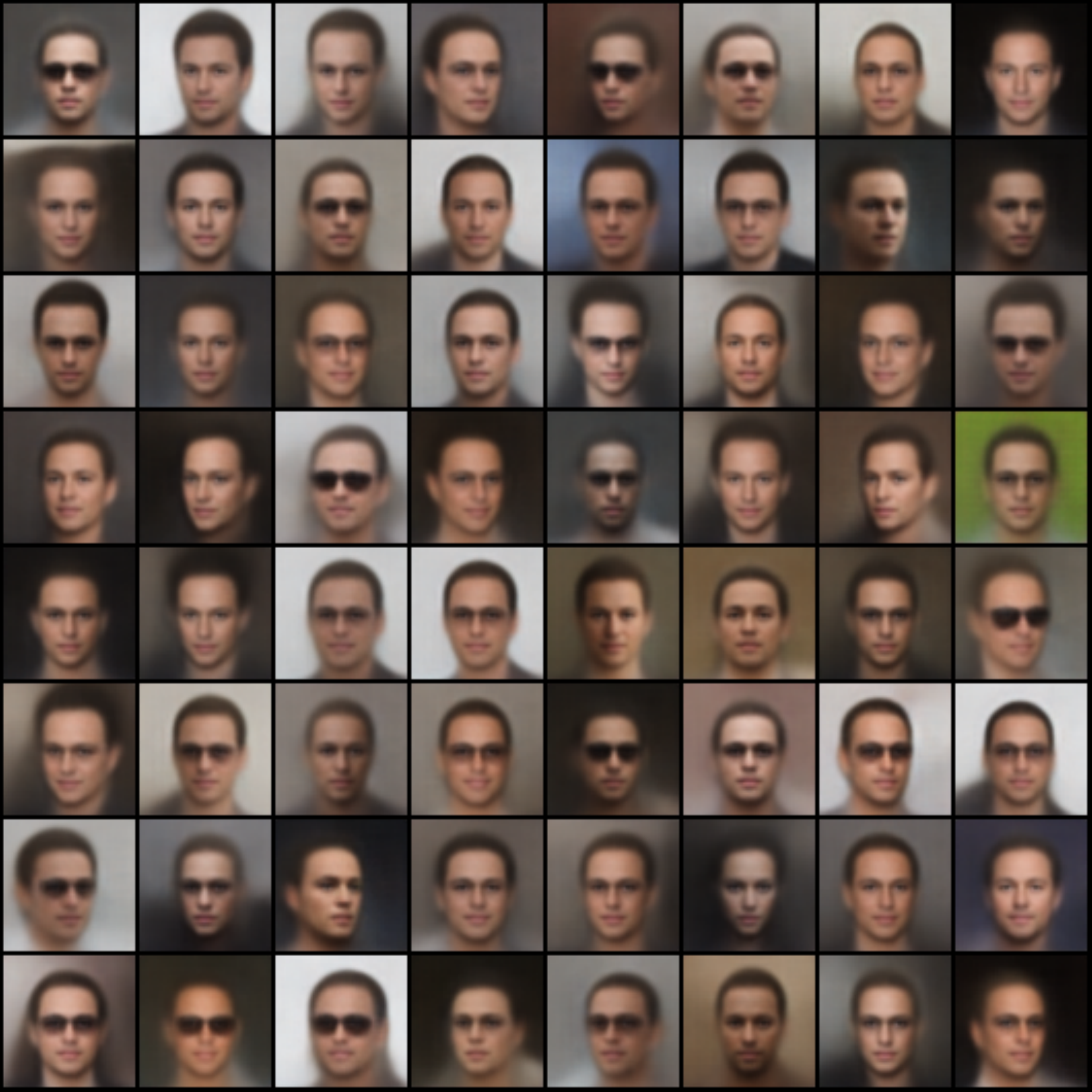}
        \caption{\textbf{MUSE (Ours)}}
        \label{fig:celeb_MUSE}
    \end{subfigure}%
    \hfill 
    \begin{subfigure}[b]{0.24\textwidth}
        \centering
        \includegraphics[height=4cm]{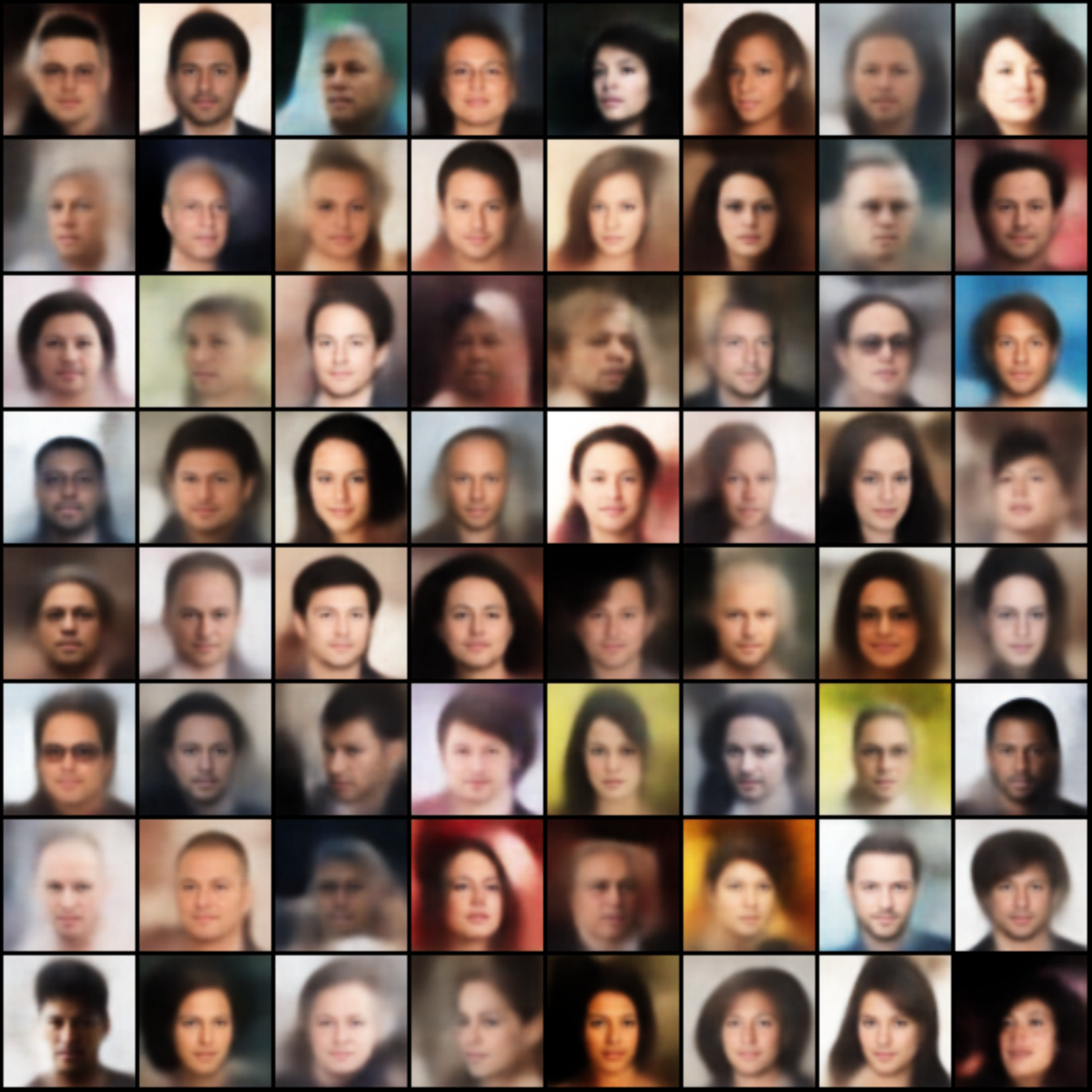}
        \caption{MVAE}
        \label{fig:celeb_mvae}
    \end{subfigure}
    \caption{Cross-modality image generation from label information: in the MNIST dataset considering ${\mathbf{x}_2 = \{2, 4, 7, 9\}}$,in the CelebA dataset considering ${\mathbf{x}_2 = \{\text{Male}, \text{Eyeglasses}, \text{Black Hair}, \text{Receding Hairline}, \text{Goatee}\}}$. The MUSE model is the only able to generate high-quality, varied and coherent image samples from label information. Best viewed with zoom.}
    \label{fig:standard_cmi_samples}
    \vspace{-2ex}
\end{figure*}

In the three-modality MNIST-SVHN scenario, MUSE, in contrast with MMVAE, is able to encode information provided by two modalities ($\mathbf{x}_2$ and $\mathbf{x}_3$) for cross-modality generation, shown by the increase in performance from ${\log p(\mathbf{x}_1 \mid \mathbf{x}_2)}$ to ${\log p(\mathbf{x}_1 \mid \mathbf{x}_2, \mathbf{x}_3)}$. MUSE also outperforms all other baselines in the CelebA dataset in regards to the cross-modality metrics, ${\log p(\mathbf{x}_1 \mid \mathbf{x}_2)}$ and ${\log p(\mathbf{x}_2 \mid \mathbf{x}_1)}$. The quantitative results are aligned with the qualitative evaluation of the generated images in Fig.~\ref{fig:standard_cmi_samples}: MUSE is the only model able to generate high-quality and diverse image samples, semantically coherent with attribute information. However, in the MNIST dataset, the quantitative results for the label-to-image conditional generation, ${\log p(\mathbf{x}_1 \mid \mathbf{x}_2)}$, are at odds with the qualitative assessment of the image samples (Fig.~\ref{fig:standard_cmi_samples}): MVAE seems to outperform MUSE in terms of ${\log p(\mathbf{x}_1 \mid \mathbf{x}_2)}$, yet fails to generate coherent image samples from label information, as shown in Fig.~\ref{fig:standard_cmi_samples}. This apparent contradiction motivates the need for the development of more suitable metrics to evaluate the performance of generative models when provided with missing modality information.

\section{Related Work}
\label{Section:RW}

Reinforcement Learning (RL) agents often employ representation models to learn to act in their equally often complex environments. Such representations can encompass the structure of the environment (state representations) \citep{lesort2018state} or of the actions of the agent (action representations) \cite{chandak2019learning}. Low-dimensional representations of the environment allow agents to learn efficiently how to perform tasks. Initial works considered state aggregation algorithms employing bisimulation metrics to provide sample-efficient learning \cite{ferns2004metrics,givan2003equivalence,li2006towards}. However, such methods are not easily scalable to scenarios with large-scale state spaces or complex dynamics. Variational autoencoders were first employed by Higgins \textit{et al.} to learn a low-dimensional representation of observations provided by the environment that allows for zero-shot policy transfer to a target domain \cite{higgins2017darla}. Similarly, VAEs are employed as a representation model in the \emph{World Model}~\cite{ha2018world} and \emph{Dreamer}~\cite{hafner2019dream} frameworks, learning to encode a low-dimensional code for image observations of the agent, at each time-step. However, all the previously discussed works assume that the agent is provided only, and always, with image observations.

Recently, the problem of transferring policies across different perceptual modalities has been proposed and addressed for Atari Games with two modalities~\cite{silva2020playing}. However, such work assumes that the agent only has access to a single modality during training and task execution. In this work, we uplift such restriction and show that agents using MUSE as a sensory representation model are able to learn policies considering joint-modality observations and directly reuse such policies, without further training, in scenarios of missing modality information.
\newline
\newline
In order to learn representations of multimodal data, multimodal VAE (MVAE) models are widely employed, extending the original model~\cite{kingma2013auto}. Early approaches, such as the AVAE and JMVAE models, considered forcing the individual representations encoded from each modality to match~\cite{suzuki2016joint,yin2017associate}. However, these models require individual neural-networks for each modality and each combination of modalities, limiting their applicability in scenarios with more than two modalities~\citep{korthals2019multi}. Other approaches proposed the factorization of the multimodal representation into separate, independent, representations, such as the MFM model ~\citep{tsai2019learning,hsu2018disentangling}. However, such models require explicit semantic (label) information to encode the multimodal representation, thus unable to perform cross-modal generation considering all possible sets of available modalities.

Recently, two approaches were proposed to learn a scalable multimodal representation, differing on the function responsible for merging multimodal information: the Multimodal VAE (MVAE)~\cite{wu2018multimodal}, employing a Product-of-Experts (PoE) solution to merge modality-specific information, and the Mixture-of-Experts MVAE (MMVAE), which employs a Mixture-of-Experts (MoE) solution to encode the multimodal representation~\cite{shi2019variational}. However, the MVAE model struggles to perform cross-modal generation robustly across all modalities due to the overconfident experts issue, neglecting information from lower-dimensional modalities, as shown in~\cite{shi2019variational}. On the other hand, the MMVAE is computationally less efficient, relying on importance sampling training schemes and $M^2$ passes through the decoder networks to train the MoE encoder, hindering its application in scenarios with arbitrary large number of modalities. Moreover, the model assumes that the modalities are of comparable complexity and, due to the MoE sampling method, is unable to merge information from multiple modalities for downstream tasks.

We summarize the main differences that distinguish our work. MUSE is simultaneously: \emph{(i)} able to scale to a large number of input modalities without requiring combinatorial number of networks, unlike latent-representation approximation methods~\citep{yin2017associate,suzuki2016joint,korthals2019multi}; \emph{(ii)} agnostic to the nature of the provided modalities, unlike factorized approaches~\citep{tsai2019learning,hsu2018disentangling} and MMVAE~\citep{shi2019variational}; \emph{ (iii)} able to robustly perform cross-modality inference regardless of the complexity of the target modality, unlike MVAE~\citep{wu2018multimodal}; \emph{(iv)} is computationally efficient to train, unlike MMVAE~\citep{shi2019variational}.

\section{Conclusion}
\label{Section:Conclusion}

We addressed the question of \emph{sensing the world}: how to learn a multimodal representation of a RL agent's environment that allows the execution of tasks under missing modality information.  We contributed MUSE, a novel multimodal sensory representation model that considers hierarchical representation spaces. Our results show that, with MUSE, RL agents  are able to perform tasks under such incomplete perceptions, outperforming other literature-standard designs. Moreover, we showed that the performance of MUSE scales to more complex scenarios, outperforming other state-of-the-art models. We can envision scenarios where MUSE would allow agents to act with unexpected damaged sensors (e.g. autonomous cars) or respecting privacy concerns (e.g. virtual assistants).

The introduction of hierarchy in multimodal generative models broadens the design possibilities of such models. In future work, we will exploit the modularity arising from having modality-specific representations, exploring the use of pretrained representation models. Additionally, we will further explore evaluation schemes to assess the quality of multimodal representations beyond standard likelihood metrics, such as following~\citet{poklukar2021geomca}. Finally, we will also explore how to provide agents with robustness to noisy observations and develop mechanisms to reason about their confidence on the information provided by different sensors.

\section{Acknowledgements}

This work was partially supported by national funds through the Portuguese Fundação para a Ciência e a Tecnologia under project UIDB/50021/2020 (INESC-ID multi annual funding) and project PTDC/CCI-COM/5060/2021. In addition, this research was partially supported by TAILOR, a project funded by EU
Horizon 2020 research and innovation programme under GA No. 952215. This work was also supported by funds from Europe Research Council under project BIRD 884887. The first author acknowledges the Fundação para a Ciência e a Tecnologia PhD grant SFRH/BD/139362/2018.

\balance
\bibliography{references}

\begin{thebibliography}{33}
\providecommand{\natexlab}[1]{#1}
\providecommand{\url}[1]{\texttt{#1}}
\expandafter\ifx\csname urlstyle\endcsname\relax
  \providecommand{\doi}[1]{doi: #1}\else
  \providecommand{\doi}{doi: \begingroup \urlstyle{rm}\Url}\fi

\bibitem[Arulkumaran et~al.(2017)Arulkumaran, Deisenroth, Brundage, and
  Bharath]{arulkumaran2017brief}
Kai Arulkumaran, Marc~Peter Deisenroth, Miles Brundage, and Anil~Anthony
  Bharath.
\newblock A brief survey of deep reinforcement learning.
\newblock \emph{arXiv preprint arXiv:1708.05866}, 2017.

\bibitem[Brockman et~al.(2016)Brockman, Cheung, Pettersson, Schneider,
  Schulman, Tang, and Zaremba]{brockman2016openai}
Greg Brockman, Vicki Cheung, Ludwig Pettersson, Jonas Schneider, John Schulman,
  Jie Tang, and Wojciech Zaremba.
\newblock Openai gym.
\newblock \emph{arXiv preprint arXiv:1606.01540}, 2016.

\bibitem[Chandak et~al.(2019)Chandak, Theocharous, Kostas, Jordan, and
  Thomas]{chandak2019learning}
Yash Chandak, Georgios Theocharous, James Kostas, Scott Jordan, and Philip
  Thomas.
\newblock Learning action representations for reinforcement learning.
\newblock In \emph{International Conference on Machine Learning}, pages
  941--950. PMLR, 2019.

\bibitem[Collier et~al.(2020)Collier, Nazabal, and Williams]{collier2020vaes}
Mark Collier, Alfredo Nazabal, and Christopher~KI Williams.
\newblock Vaes in the presence of missing data.
\newblock \emph{arXiv preprint arXiv:2006.05301}, 2020.

\bibitem[Ferns et~al.(2004)Ferns, Panangaden, and Precup]{ferns2004metrics}
Norm Ferns, Prakash Panangaden, and Doina Precup.
\newblock Metrics for finite markov decision processes.
\newblock In \emph{UAI}, volume~4, pages 162--169, 2004.

\bibitem[Gelada et~al.(2019)Gelada, Kumar, Buckman, Nachum, and
  Bellemare]{gelada2019deepmdp}
Carles Gelada, Saurabh Kumar, Jacob Buckman, Ofir Nachum, and Marc~G Bellemare.
\newblock Deepmdp: Learning continuous latent space models for representation
  learning.
\newblock In \emph{International Conference on Machine Learning}, pages
  2170--2179, 2019.

\bibitem[Givan et~al.(2003)Givan, Dean, and Greig]{givan2003equivalence}
Robert Givan, Thomas Dean, and Matthew Greig.
\newblock Equivalence notions and model minimization in markov decision
  processes.
\newblock \emph{Artificial Intelligence}, 147\penalty0 (1-2):\penalty0
  163--223, 2003.

\bibitem[Ha and Schmidhuber(2018)]{ha2018world}
David Ha and J{\"u}rgen Schmidhuber.
\newblock World models.
\newblock \emph{arXiv preprint arXiv:1803.10122}, 2018.

\bibitem[Hafner et~al.(2019)Hafner, Lillicrap, Ba, and
  Norouzi]{hafner2019dream}
Danijar Hafner, Timothy Lillicrap, Jimmy Ba, and Mohammad Norouzi.
\newblock Dream to control: Learning behaviors by latent imagination.
\newblock \emph{arXiv preprint arXiv:1912.01603}, 2019.

\bibitem[Havtorn et~al.(2021)Havtorn, Frellsen, Hauberg, and
  Maal{\o}e]{havtorn2021hierarchical}
Jakob~D Havtorn, Jes Frellsen, S{\o}ren Hauberg, and Lars Maal{\o}e.
\newblock Hierarchical vaes know what they don't know.
\newblock \emph{arXiv preprint arXiv:2102.08248}, 2021.

\bibitem[Higgins et~al.(2017)Higgins, Pal, Rusu, Matthey, Burgess, Pritzel,
  Botvinick, Blundell, and Lerchner]{higgins2017darla}
Irina Higgins, Arka Pal, Andrei Rusu, Loic Matthey, Christopher Burgess,
  Alexander Pritzel, Matthew Botvinick, Charles Blundell, and Alexander
  Lerchner.
\newblock Darla: Improving zero-shot transfer in reinforcement learning.
\newblock In \emph{Proceedings of the 34th International Conference on Machine
  Learning-Volume 70}, pages 1480--1490. JMLR. org, 2017.

\bibitem[Hsu and Glass(2018)]{hsu2018disentangling}
Wei-Ning Hsu and James Glass.
\newblock Disentangling by partitioning: A representation learning framework
  for multimodal sensory data.
\newblock \emph{arXiv preprint arXiv:1805.11264}, 2018.

\bibitem[Kingma and Welling(2013)]{kingma2013auto}
Diederik~P Kingma and Max Welling.
\newblock Auto-encoding variational bayes.
\newblock \emph{arXiv preprint arXiv:1312.6114}, 2013.

\bibitem[Korthals et~al.(2019)Korthals, Rudolph, Leitner, Hesse, and
  R{\"u}ckert]{korthals2019multi}
Timo Korthals, Daniel Rudolph, J{\"u}rgen Leitner, Marc Hesse, and Ulrich
  R{\"u}ckert.
\newblock Multi-modal generative models for learning epistemic active sensing.
\newblock In \emph{2019 IEEE International Conference on Robotics and
  Automation}, 2019.

\bibitem[LeCun et~al.(1998)LeCun, Bottou, Bengio, and
  Haffner]{lecun1998gradient}
Yann LeCun, L{\'e}on Bottou, Yoshua Bengio, and Patrick Haffner.
\newblock Gradient-based learning applied to document recognition.
\newblock \emph{Proceedings of the IEEE}, 86\penalty0 (11):\penalty0
  2278--2324, 1998.

\bibitem[Lesort et~al.(2018)Lesort, D{\'\i}az-Rodr{\'\i}guez, Goudou, and
  Filliat]{lesort2018state}
Timoth{\'e}e Lesort, Natalia D{\'\i}az-Rodr{\'\i}guez, Jean-Franois Goudou, and
  David Filliat.
\newblock State representation learning for control: An overview.
\newblock \emph{Neural Networks}, 108:\penalty0 379--392, 2018.

\bibitem[Li et~al.(2006)Li, Walsh, and Littman]{li2006towards}
Lihong Li, Thomas~J Walsh, and Michael~L Littman.
\newblock Towards a unified theory of state abstraction for mdps.
\newblock \emph{ISAIM}, 4:\penalty0 5, 2006.

\bibitem[Lillicrap et~al.(2015)Lillicrap, Hunt, Pritzel, Heess, Erez, Tassa,
  Silver, and Wierstra]{lillicrap2015continuous}
Timothy~P Lillicrap, Jonathan~J Hunt, Alexander Pritzel, Nicolas Heess, Tom
  Erez, Yuval Tassa, David Silver, and Daan Wierstra.
\newblock Continuous control with deep reinforcement learning.
\newblock \emph{arXiv preprint arXiv:1509.02971}, 2015.

\bibitem[Liu et~al.(2015)Liu, Luo, Wang, and Tang]{liu2015faceattributes}
Ziwei Liu, Ping Luo, Xiaogang Wang, and Xiaoou Tang.
\newblock Deep learning face attributes in the wild.
\newblock In \emph{Proceedings of International Conference on Computer Vision
  (ICCV)}, December 2015.

\bibitem[Mnih et~al.(2015)Mnih, Kavukcuoglu, Silver, Rusu, Veness, Bellemare,
  Graves, Riedmiller, Fidjeland, Ostrovski, et~al.]{mnih2015human}
Volodymyr Mnih, Koray Kavukcuoglu, David Silver, Andrei~A Rusu, Joel Veness,
  Marc~G Bellemare, Alex Graves, Martin Riedmiller, Andreas~K Fidjeland, Georg
  Ostrovski, et~al.
\newblock Human-level control through deep reinforcement learning.
\newblock \emph{nature}, 518\penalty0 (7540):\penalty0 529--533, 2015.

\bibitem[Netzer et~al.(2011)Netzer, Wang, Coates, Bissacco, Wu, and
  Ng]{netzer2011reading}
Yuval Netzer, Tao Wang, Adam Coates, Alessandro Bissacco, Bo~Wu, and Andrew~Y
  Ng.
\newblock Reading digits in natural images with unsupervised feature learning.
\newblock 2011.

\bibitem[Partan(2017)]{partan2017multimodal}
Sarah~R Partan.
\newblock Multimodal shifts in noise: switching channels to communicate through
  rapid environmental change.
\newblock \emph{Animal Behaviour}, 124:\penalty0 325--337, 2017.

\bibitem[Poklukar et~al.(2021)Poklukar, Varava, and Kragic]{poklukar2021geomca}
Petra Poklukar, Anastasia Varava, and Danica Kragic.
\newblock Geomca: Geometric evaluation of data representations.
\newblock \emph{arXiv preprint arXiv:2105.12486}, 2021.

\bibitem[Shi et~al.(2019)Shi, Siddharth, Paige, and Torr]{shi2019variational}
Yuge Shi, N~Siddharth, Brooks Paige, and Philip Torr.
\newblock Variational mixture-of-experts autoencoders for multi-modal deep
  generative models.
\newblock In \emph{Advances in Neural Information Processing Systems}, pages
  15692--15703, 2019.

\bibitem[Silva et~al.(2020)Silva, Vasco, Melo, Paiva, and
  Veloso]{silva2020playing}
Rui Silva, Miguel Vasco, Francisco~S Melo, Ana Paiva, and Manuela Veloso.
\newblock Playing games in the dark: An approach for cross-modality transfer in
  reinforcement learning.
\newblock In \emph{Proceedings of the 19th International Conference on
  Autonomous Agents and MultiAgent Systems}, pages 1260--1268, 2020.

\bibitem[Sutton and Barto(1998)]{sutton1998reinforcement}
Richard Sutton and Andrew Barto.
\newblock \emph{{Reinforcement Learning: An Introduction}}.
\newblock MIT press Cambridge, 1998.

\bibitem[Suzuki et~al.(2016)Suzuki, Nakayama, and Matsuo]{suzuki2016joint}
Masahiro Suzuki, Kotaro Nakayama, and Yutaka Matsuo.
\newblock Joint multimodal learning with deep generative models.
\newblock \emph{arXiv preprint arXiv:1611.01891}, 2016.

\bibitem[Tsai et~al.(2019)Tsai, Liang, Zadeh, Morency, and
  Salakhutdinov]{tsai2019learning}
Yao-Hung~Hubert Tsai, Paul~Pu Liang, Amir Zadeh, Louis-Philippe Morency, and
  Ruslan Salakhutdinov.
\newblock Learning factorized multimodal representations.
\newblock In \emph{International Conference on Representation Learning}, 2019.

\bibitem[Vecerik et~al.(2017)Vecerik, Hester, Scholz, Wang, Pietquin, Piot,
  Heess, Roth{\"o}rl, Lampe, and Riedmiller]{vecerik2017leveraging}
Mel Vecerik, Todd Hester, Jonathan Scholz, Fumin Wang, Olivier Pietquin, Bilal
  Piot, Nicolas Heess, Thomas Roth{\"o}rl, Thomas Lampe, and Martin Riedmiller.
\newblock Leveraging demonstrations for deep reinforcement learning on robotics
  problems with sparse rewards.
\newblock \emph{arXiv preprint arXiv:1707.08817}, 2017.

\bibitem[Watkins(1989)]{watkins89phd}
Christopher Watkins.
\newblock \emph{{Learning from delayed rewards}}.
\newblock PhD thesis, Cambridge University, 1989.

\bibitem[Wu and Goodman(2018)]{wu2018multimodal}
Mike Wu and Noah Goodman.
\newblock Multimodal generative models for scalable weakly-supervised learning.
\newblock In \emph{Advances in Neural Information Processing Systems}, pages
  5575--5585, 2018.

\bibitem[Yin et~al.(2017)Yin, Melo, Billard, and Paiva]{yin2017associate}
Hang Yin, Francisco~S Melo, Aude Billard, and Ana Paiva.
\newblock Associate latent encodings in learning from demonstrations.
\newblock In \emph{Thirty-First AAAI Conference on Artificial Intelligence},
  2017.

\bibitem[Zhang et~al.(2019)Zhang, Vikram, Smith, Abbeel, Johnson, and
  Levine]{zhang2019solar}
Marvin Zhang, Sharad Vikram, Laura Smith, Pieter Abbeel, Matthew Johnson, and
  Sergey Levine.
\newblock Solar: Deep structured representations for model-based reinforcement
  learning.
\newblock In \emph{International Conference on Machine Learning}, pages
  7444--7453. PMLR, 2019.

\end{thebibliography}
\bibliographystyle{plainnat}

\clearpage
\onecolumn

\appendix

\section{Ablation Study}

\begin{figure*}[t]
    \centering
    \begin{subfigure}[b]{0.31\textwidth}
        \centering
        \includegraphics[height=3.4cm]{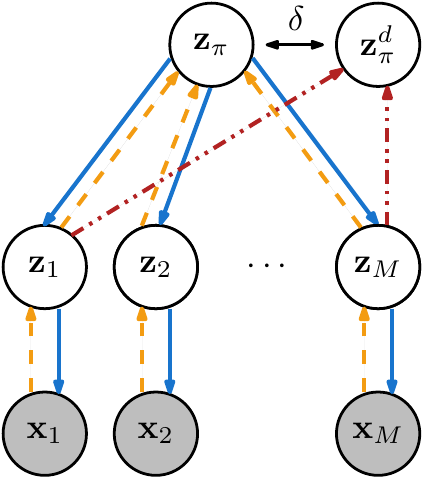}
        \caption{MUSE}
        \label{fig:ablation:model}
    \end{subfigure}%
    \hfill 
    \begin{subfigure}[b]{0.31\textwidth}
        \centering
         \includegraphics[height=3.4cm]{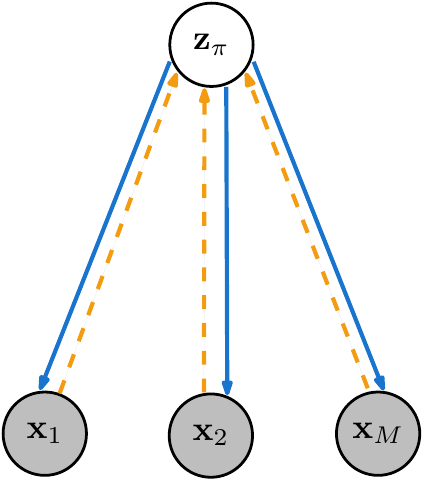}
        \caption{MUSE$_{\text{H}}$}
        \label{fig:ablation:h}
    \end{subfigure}
    \hfill 
      \begin{subfigure}[b]{0.31\textwidth}
        \centering
        \includegraphics[height=3.4cm]{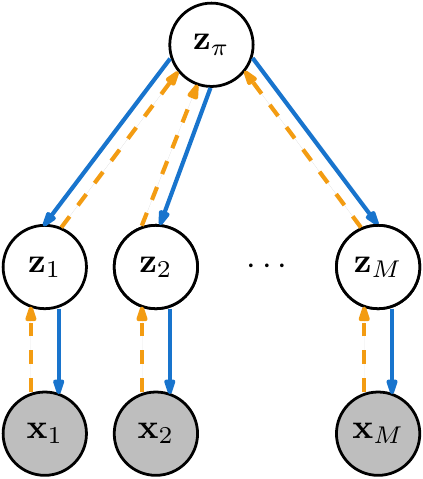}
        \caption{MUSE$_{\text{A}}$}
        \label{fig:ablation:a}
    \end{subfigure}
    \caption{The ablated versions of the MUSE model (a), considered for this study: (b) MUSE$_{\text{H}}$, a non-hierarchical version of MUSE; (c) MUSE$_{\text{A}}$, a version trained without resorting to the ALMA training scheme.}
    \label{fig:ablation}
    \vspace{-3ex}
\end{figure*}

We present an ablation study to evaluate the role of each component of MUSE in the generation of high-quality, coherent samples through cross-modal inference. To do so, we instantiate three different versions of MUSE, as depicted in Fig.~\ref{fig:ablation}:
\begin{itemize}
    \item MUSE$\phantom{_{\text{H}}}$ - The standard version of MUSE (Fig.~\ref{fig:ablation:model});
    \item MUSE$_{\text{H}}$ - A \emph{non-hierarchical} version of MUSE (Fig.~\ref{fig:ablation:h});
    \item MUSE$_{\text{A}}$ - A hierarchical version of MUSE that employs a \emph{regular PoE solution}, without the proposed ALMA training scheme (Fig.~\ref{fig:ablation:a}).
\end{itemize}

\begin{figure*}[t]
    \centering
    \begin{subfigure}[b]{0.31\textwidth}
        \centering
        \includegraphics[height=4.3cm]{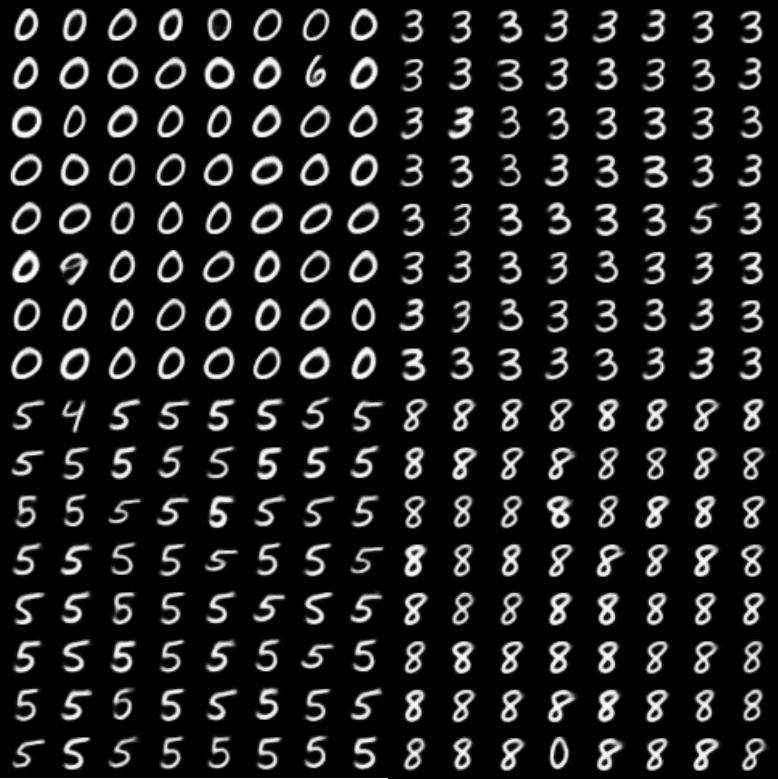}
        \caption{\textbf{MUSE}}
        \label{fig:appendix_mnist_muse}
    \end{subfigure}%
    \hfill 
    \begin{subfigure}[b]{0.31\textwidth}
        \centering
        \includegraphics[height=4.3cm]{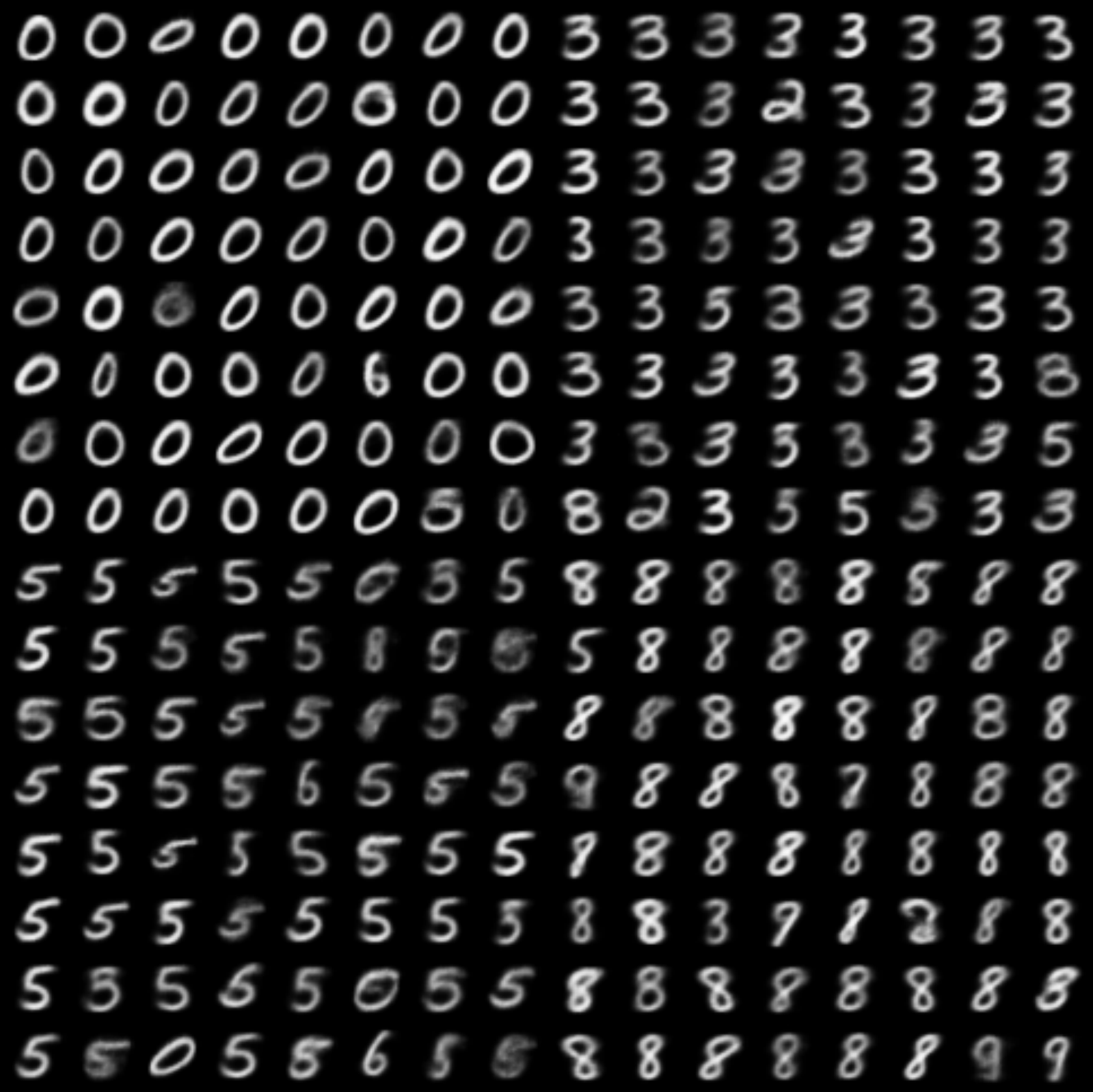}
        \caption{MUSE$_{\text{H}}$}
        \label{fig:appendix_mnist_muse_h}
    \end{subfigure}
    \hfill 
    \begin{subfigure}[b]{0.31\textwidth}
        \centering
        \includegraphics[height=4.3cm]{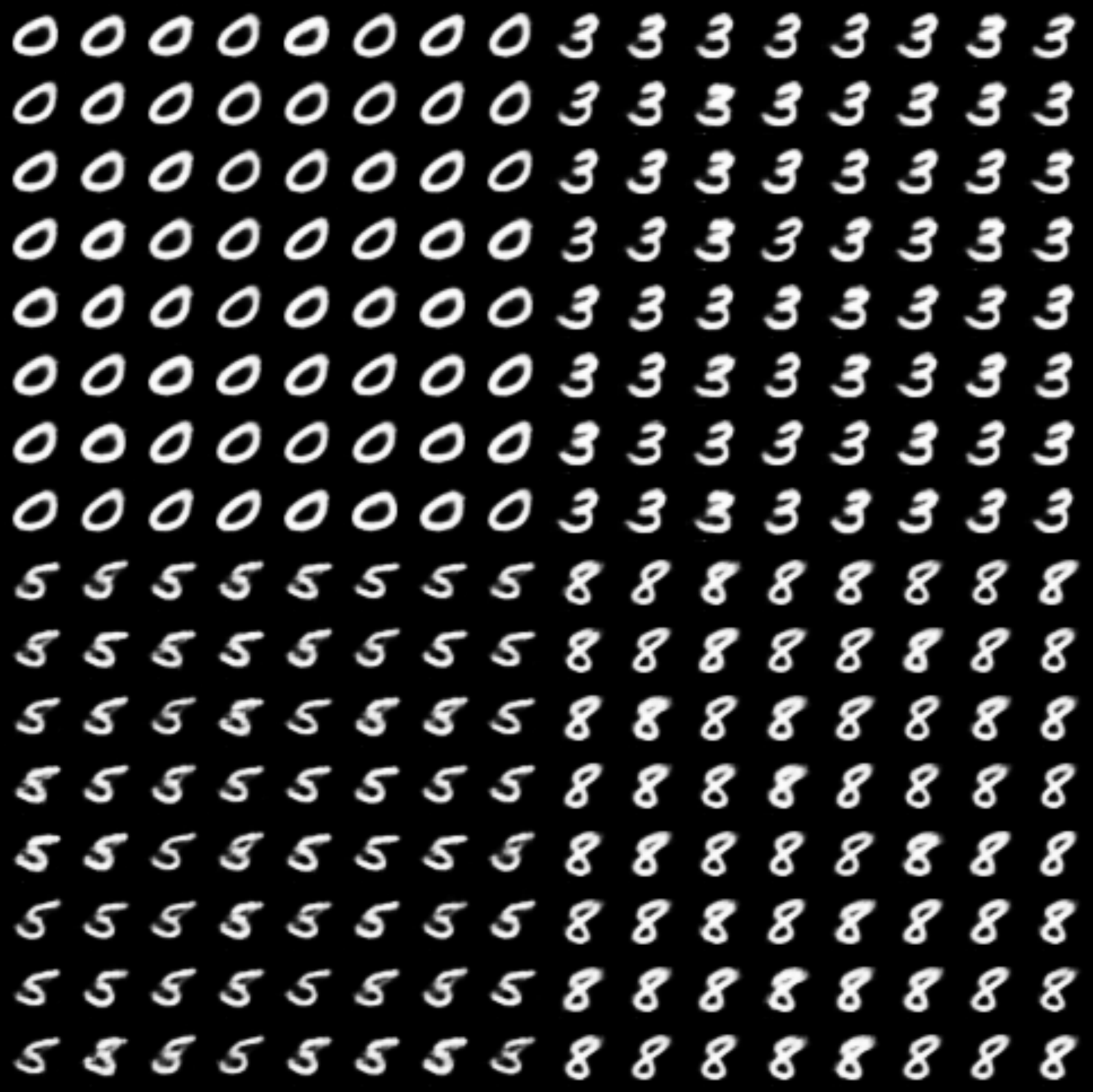}
        \caption{MUSE$_{\text{A}}$}
        \label{fig:appendix_mnist_muse_a}
    \end{subfigure}
    \caption{Cross-modal image generation from label information in the MNIST dataset considering ${\mathbf{x}_2 = \{0, 3, 5, 8\}}$ (Best viewed with zoom).}
    \label{fig:ablation:samples}
\end{figure*}

The MUSE$_{\text{H}}$ ablated version allows the evaluation of the role of the hierarchical representation spaces for the performance of the model. The MUSE$_{\text{A}}$ ablated version allows the evaluation of the ALMA training scheme in the generation of coherent, high-quality samples. We evaluate all ablated models in the MNIST dataset considering the standard log-likelihood metrics. We present the evaluation results in Table~\ref{Table:ablation_results}, estimated resorting to 5000 importance-weighted samples and averaged over 5 independent runs\footnote{The results for the MUSE$_{\text{A}}$ version are averaged over 3 independent runs, as the training of the remaining two runs diverged, hinting at the unstable training of this version.}. We present image samples generated by each model from label information in Fig.~\ref{fig:ablation:samples}.

The results in Table~\ref{Table:ablation_results} attest the role of the hierarchical representation spaces in the overall performance of MUSE: the original MUSE model outperforms the non-hierarchical version (MUSE$_{\text{H}})$ in marginal likelihood $\left(\log p(\mathbf{x}_1), \log p(\mathbf{x}_2)\right)$, learning a richer modality-specific representation than the non-hierarchical version. While the non-hierarchical version outperforms MUSE on joint and conditional likelihoods, visual inspection of the image samples presented in Fig.~\ref{fig:appendix_mnist_muse_h} show that MUSE$_{\text{H}}$ is unable to generate high-quality samples. Once again, the results attest the importance of considering modality-specific representation spaces that allow the model to generate high-quality sample, regardless of the complexity of the modality.

The results in Table~\ref{Table:ablation_results} also attest the fundamental role of the ALMA training for the performance of MUSE: the original MUSE model outperforms the MUSE$_{\text{A}}$ version in joint and conditional likelihood. The original PoE solution employed by MUSE$_{\text{A}}$ struggles to generate coherent information for all modalities, as shown by the result of cross-modal accuracy in Table~\ref{Table:ablation_results}. This results hints that the model is suffering from the overconfident expert problem discussed in 
~\citet{shi2019variational}.

\begin{table}[b]
\centering
\caption{Metrics for generative performance in the MNIST dataset of the ablated versions of the model (best results in bold). All results averaged over 5 independent runs, considering 5000 importance samples.}
\begin{subtable}{\textwidth}
\centering
\caption{Standard Metrics}
\begin{adjustbox}{width=0.73\columnwidth,center}
\begin{tabular}{@{}lccccc@{}}
\toprule
 Model & $\,\,\,\log p(\mathbf{x}_1)$ & $\,\,\,\log p(\mathbf{x}_2)$ & $\,\,\,\log p(\mathbf{x}_1,  \mathbf{x}_2)$ & $\,\,\,\log p(\mathbf{x}_1 \mid \mathbf{x}_2)$ & $\,\,\,\log p(\mathbf{x}_2 \mid \mathbf{x}_1)$ \\ \midrule
\textbf{MUSE} & $\bf-24.06 \pm 0.03$  & $ -2.41 \pm 0.02$  & $-34.72 \pm 1.47$  & $-33.97 \pm 1.16$   & $ -\phantom{3}4.72 \pm 0.15$  \\\midrule
MUSE$_{\text{H}}$ & $-27.91 \pm 0.16$  & $\bf-2.31 \pm 0.01$  & $\bf -28.09 \pm 0.17$  & $\bf -28.49 \pm 0.10$  & $\bf-\phantom{3}2.90 \pm 0.02$  \\
MUSE$_{\text{A}}$ & $\bf-24.06 \pm 0.02$  & $-2.39 \pm 0.01$  & $ -59.63 \pm 0.78$  & $ -80.71 \pm 7.15$  & $-39.00 \pm 9.93$  \\ \bottomrule
\end{tabular}
\end{adjustbox}
\end{subtable}
\label{Table:ablation_results}
\end{table}

\section{Description of Multimodal Atari Games}

\begin{figure}[t]
\centering
\begin{subfigure}{0.48\columnwidth}
        \centering
        \includegraphics[height=4cm, trim=0.0cm 0.0cm 0.0cm 0.0cm]{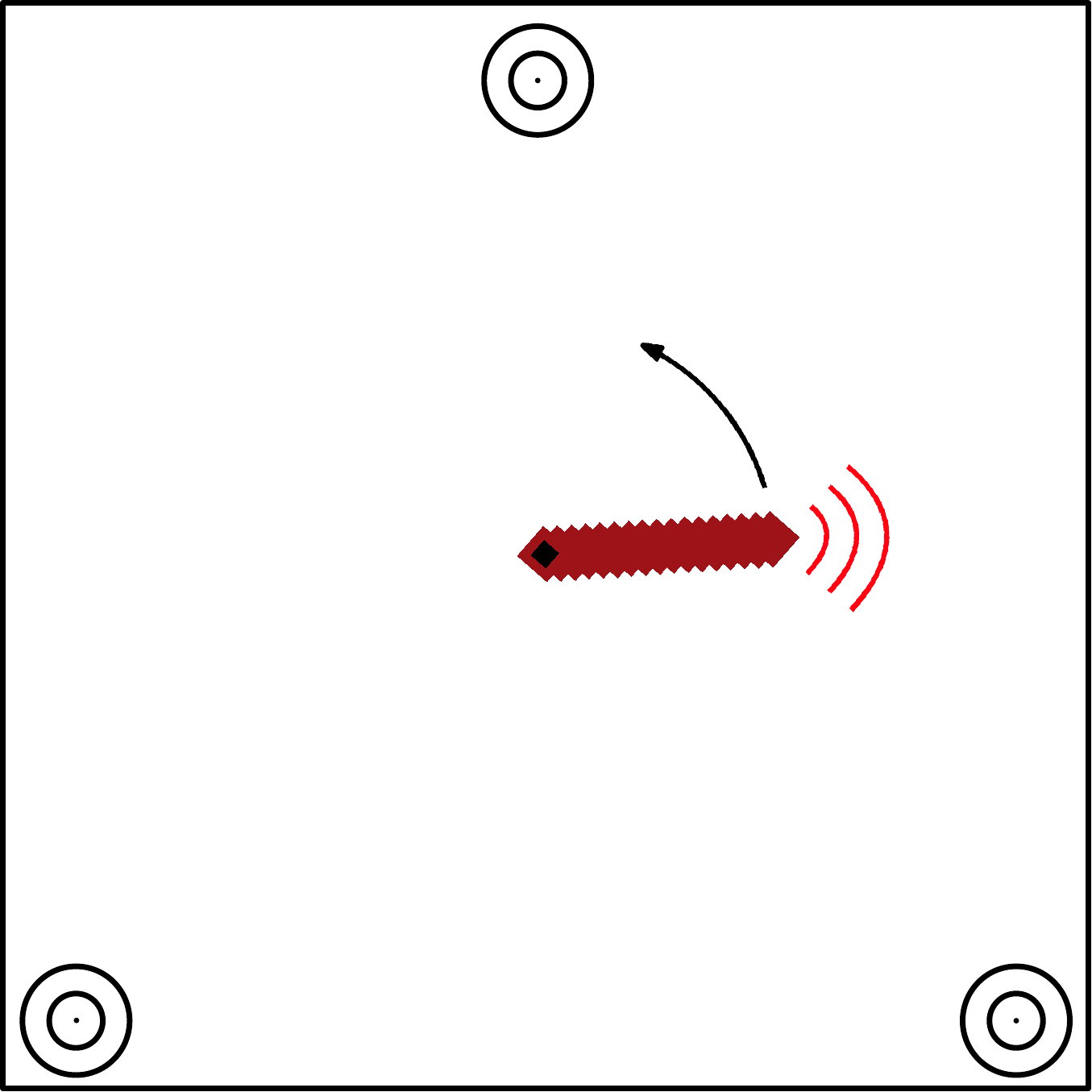}
        \caption{Pendulum}
        \label{fig:atari_pendulum}
    \end{subfigure}%
    \hfill 
    \begin{subfigure}{0.48\columnwidth}
        \centering
        \includegraphics[height=4cm, trim=0.0cm 0.0cm 0.0cm 0.0cm]{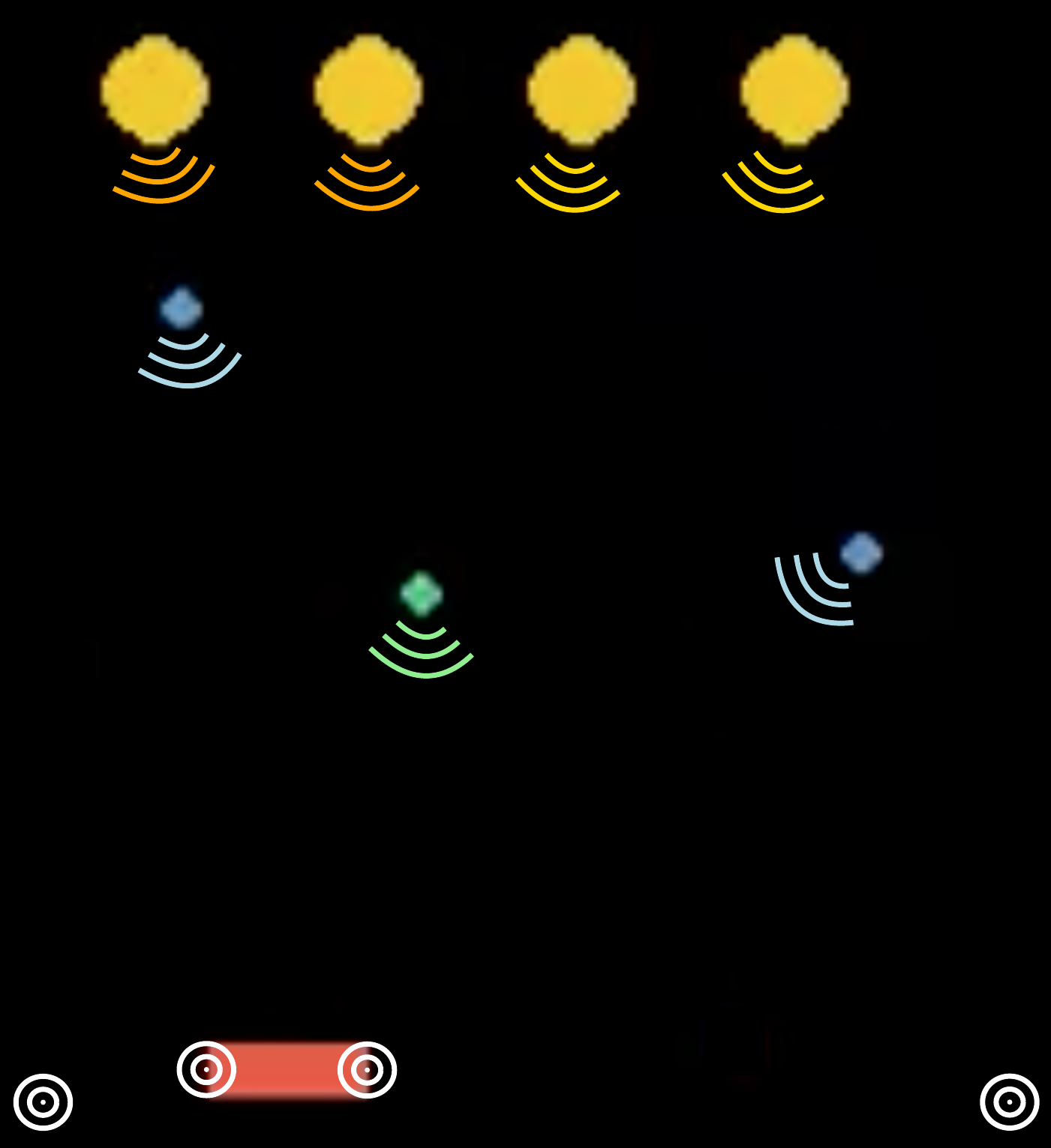}
        \caption{\textsc{Hyperhot}}
        \label{fig:atari_hyperhot}
    \end{subfigure}
\caption{\emph{Multimodal} Atari Games, depicting the image and sound perceptual information provided to the agent, along with the sound receivers of the agent (concentric circles).}
\label{fig:atari_games}
\vspace{-2ex}
\end{figure}

We provide a description of the \emph{multimodal Atari games} scenarios, employed in Section~\ref{Section:eval:rl} and first introduced in~\citet{silva2020playing}. Contrary to the standard Atari games, in which the agent only receives visual information from the game environment, multimodal Atari games allows the agent to receive information from the environment through additional modality channels. In Table~\ref{Table:atari_games} we present the numerical results of the comparative study presented in Section~\ref{Section:eval:rl}.

\subsection{\textsc{pendulum} Environment}

The inverted \textsc{pendulum} environment is a classic control problem, where the goal is to swing the pendulum up so it stays upright. The multimodal pendulum environment, depicted in Figure~\ref{fig:atari_pendulum}, is a modified version from OpenAI gym that includes both an image and a sound component as the observations of the environment.

The sound component is generated by the tip of the pendulum, emitting a constant frequency $f_0$. This frequency is received by a set of $S$ sound receivers $\left\lbrace \rho_1, \dots, \rho_S \right\rbrace$. At each timestep, the frequency $f'_i$ heard by each sound receiver $\rho_i$ is modified by the Doppler effect, modifying the frequency heard by an observer as a function of the velocity of the sound emitter,
\begin{equation*}
  f'_i
  =
  \left(
    \frac
    {c + \dot{\rho}_i \cdot \left( e - \rho_i \right)}
    {c - \dot{e} \cdot \left( \rho_i - e \right)}
  \right)
  f_0,
\end{equation*}
where $\rho_i$ is the position of sound observer, $e$ is the position of the sound emitter, and the dot notation represent the velocities. Moreover, $c$ is the speed of sound in the environment. In addition to the frequency, the scenario also accounts for the change in amplitude as a function of the relative position of the emitter in relation to the observer: the amplitude $a_i$ heard by receiver $\rho_i$ follows the inverse square law,
\begin{equation*}
  a_i = \frac{K}{\|\vec{e} - \vec{\rho}_i\|^2},
\end{equation*}
where $K$ is a scaling constant.  In this scenario, we employ a multimodal representation space $\mathbf{z}_{\pi} \in \mathbb{R}^{10}$ for all models. For MUSE, we set the image-specific latent space $\mathbf{z}_1 \in \mathbb{R}^{16}$, the sound-specific latent space $\mathbf{z}_2 \in \mathbb{R}^{8}$;

\begin{table*}[b]
    \centering
    \caption{Performance after zero-shot policy transfer in (a) the Pendulum scenario; and (b) the \textsc{Hyperhot} scenario. At test time, the agent is provided with either image ($\mathbf{x}_\text{I}$), sound ($\mathbf{x}_\text{S}$), or joint-modality ($\mathbf{x}_\text{I}, \mathbf{x}_\text{S}$) observations. All results averaged over 10 randomly seeded runs. Higher is better.}
    \centering
   \begin{subtable}{\textwidth}
\centering
\caption{Pendulum}
\begin{adjustbox}{width=0.95\textwidth,center}
\begin{tabular}{@{}lcccccc@{}}
\toprule
       Observation & Multimodal DDPG & Multimodal DDPG (D)  & VAE + DDPG & MVAE + DDPG & \textbf{MUSE + DDPG (Ours)}     \\ \midrule
Joint $(\mathbf{x}_\text{I}$, $\mathbf{x}_\text{S})$ & $-1.292 \pm 0.439$ & $-1.157 \pm 0.207$ & $-1.086 \pm 0.141$ & $-1.030 \pm 0.092$ & $ -0.933 \pm 0.077$ \\
Image $(\mathbf{x}_\text{I})$ & $-1.279 \pm 0.412$ &  $-1.280 \pm 0.424$ &  $ -1.172 \pm 0.141$  & $-1.201 \pm 0.387$ & $-1.473 \pm 0.642$ \\
Sound $(\mathbf{x}_\text{S})$ & $-6.172 \pm 0.240$ & $-5.974 \pm 0.769$ & $-6.543 \pm 0.170$ & $-5.153 \pm 0.630$ & $\bf-3.302 \pm 0.572$ \\ \bottomrule
\end{tabular}
\end{adjustbox}
\label{Table:atari_games:pendulum}
\end{subtable}

\vspace{1ex}

\begin{subtable}{\textwidth}
\centering
\caption{\textsc{Hyperhot}}
\begin{adjustbox}{width=0.94\textwidth,center}
\begin{tabular}{@{}lccccccc@{}}
\toprule
       Observation & Multimodal DQN & Multimodal DQN (D) & VAE + DQN  & MVAE + DQN & \textbf{MUSE + DQN (Ours)}  \\ \midrule
Joint $(\mathbf{x}_\text{I}$, $\mathbf{x}_\text{S})$ & $ \phantom{-}0.910 \pm 0.196$ & $\bf \phantom{-}0.921 \pm 0.178$ & $\phantom{-}0.134 \pm 0.128$ & $\phantom{-}0.196 \pm 0.099$ & $ 0.302 \pm 0.109$  \\
Image $(\mathbf{x}_\text{I})$ & $-0.379 \pm 0.182$ & $-0.467 \pm 0.112$ & $-0.393 \pm 0.278$ & $\phantom{-}0.044 \pm 0.052$ & $\bf 0.319 \pm 0.109$  \\
Sound $(\mathbf{x}_\text{S})$ & $-0.342 \pm 0.201$ & $-0.216 \pm 0.274$ & $-0.378 \pm 0.203$  & $-0.311 \pm 0.126 $ & $\bf 0.038 \pm 0.241$  \\ \bottomrule
\end{tabular}
\end{adjustbox}
\label{Table:atari_games:hyperhot}
\end{subtable}
\label{Table:atari_games}
\end{table*}

\subsection{\textsc{Hyperhot} environment}

The \textsc{hyperhot} scenario is a top-down shooter game scenario inspired by the \emph{space invaders} Atari game, as shown in Fig.~\ref{fig:atari_hyperhot}. In this scenario, the agent receives both image and sound observations of the environment. Contrary to the pendulum scenario, in this scenario the sound is generated by multiple entities $e_i$, emitting a class-specific frequency $f_0^{(i)}$:
\begin{itemize}
\item Left-side enemy units, $e_0$ emit sounds with frequency $f_0^{(0)}$ and amplitude $a_0^{(0)}$;
\item Right-side enemy units, $e_1$ emit sounds with frequency $f_0^{(1)}$ and amplitude $a_0^{(1)}$;
\item Enemy bullets, $e_2$, emit sounds with frequency $f_0^{(2)}$ and amplitude $a_0^{(2)}$;
\item The agent's bullets, $e_3$, emit sounds with frequency $f_0^{(3)}$ and amplitude $a_0^{(3)}$.
\end{itemize}
The sounds produced by these entities are received by a set of $S$ sound receivers $\left\lbrace \rho_1, \dots, \rho_S \right\rbrace$, shown in Fig.~\ref{fig:atari_hyperhot} as the concentric circles. In this scenario the sound received by each sound receiver $\rho_j$ is modeled the sinusoidal wave of each sound-emitter $e_i$ considering its specific frequency $f_0^{(i)}$, amplitude $a_0^{(i)}$ and the distance between them, following,
\begin{equation*}
  a^{(i)}
  =
  a_0^{(i)} \exp{\left(-\delta \| e_i - \rho_j \|^2 \right)},
\end{equation*}
with $\delta$ a scaling constant and (slightly abusing the notation) $e_i$ and $\rho_j$ denote the positions of sound emitter $e_i$ and sound receiver $\rho_j$, respectively. Each sinusoidal wave is generated for a total of $1047$ discrete time steps, considering an audio sample rate of $31400$ Hz and a video frame-rate of 30 fps (similarly to what is performed in real Atari videogames). Each sound receiver sums all emitted waves and encode the amplitude values in 16-bit audio depth with amplitude in the range of $[-a_M, a_M]$, where $a_M$ is a predefined constant.

In this scenario the goal of the agent is to shoot (in green) the enemies above (in yellow), while avoiding their bullets (in blue). To do so, the agent is able to move left or right, along the bottom of the screen. The agent is rewarded for shooting the enemies, with the following reward function:
\begin{equation*}
  r =
  \begin{cases}
    10 & \text{if the agent kills all enemies, \emph{i.e.}, win}          \\
    -1 & \text{if the agent is killed or time runs out, \emph{i.e.}, lose} \\
    0  & \text{otherwise}
  \end{cases}
\end{equation*}

In this scenario, we employ a multimodal representation space $\mathbf{z}_{\pi} \in \mathbb{R}^{40}$ for all models. For MUSE, we set the image-specific latent space $\mathbf{z}_1 \in \mathbb{R}^{64}$, the sound-specific latent space $\mathbf{z}_2 \in \mathbb{R}^{64}$;

\section{Description of Evaluation Metrics}

In Section~\ref{Section:eval:generative}, we evaluate the generative performance of MUSE following standard log-likelihood metrics. We start by estimating the marginal log-likelihoods $\log p(\mathbf{x}_i)$ through importance-weighted sampling, following the standard single-modality evidence lower-bound,
\begin{equation}
\log p(\mathbf{x}_i) \leq \EX_{\,\mathbf{z}_i^{1:N} \sim q_{\phi_{i}}(\cdot \mid \mathbf{x}_{i})} \left[\log \sum_{n=1}^{N} \left( \frac{p_{\theta_{i}} (\mathbf{x}_{i}\mid \mathbf{z}_{i}^{n}) \, p(\mathbf{z}_{i}^{n})}{q_{\phi_{i}}(\mathbf{z}_i^{n} \mid \mathbf{x}_{i})}  \right)\right].    
\end{equation}
To evaluate the joint-modality log-likelihood ${\log p(\mathbf{x}_{i:M})}$, we compute a importance-weighted estimate using the standard multimodal evidence lower-bound,
\begin{equation}
\log p(\mathbf{x}_{i:M}) \leq \EX_{\,\mathbf{z}_{\pi}^{1:N} \sim q_{\phi}(\cdot \mid \mathbf{x}_{i:M})} \left[\log \sum_{n=1}^{N} \left( \frac{p(\mathbf{z}_{\pi}^{n}) \prod_{i}^{M} p_{\theta_{i}} (\mathbf{x}_{i}\mid \mathbf{z}_{\pi}^{n})}{q_{\phi}(\mathbf{z}_{\pi}^{n} \mid \mathbf{x}_{i:M})}  \right)\right],
\label{Eq:joint-likelihood}
\end{equation}
where the posterior distribution $q_{\phi}(\mathbf{z}_{\pi}^{n} \mid \mathbf{x}_{i:M}) = q_{\phi_{i:M}}(\mathbf{c}_{i:M} \mid \mathbf{x}_{i:M})\, q_{\phi_{i:M}}(\mathbf{z}_{\pi}^{n} \mid \mathbf{c}_{i:M})$ accounts for the data pass from the bottom-level decoders $q_{\phi_{i:M}}(\mathbf{c}_{i:M} \mid \mathbf{x}_{i:M})$ to the top-level decoders $q_{\phi_{i:M}}(\mathbf{z}_{\pi}^{n} \mid \mathbf{c}_{i:M})$. Finally, following \citet{shi2019variational}, to compute the conditional log-likelihood ${\log p(\mathbf{x}_{i}\mid \mathbf{x}_{\neg j})}$ we resort to the corresponding single variational posterior $q(\mathbf{z}_{\pi}\mid \mathbf{x}_{\neg i})$,
\begin{equation}
\log p(\mathbf{x}_{i}\mid \mathbf{x}_{\neg i}) \leq \EX_{\,\mathbf{z}_{\pi}^{1:N} \sim q_{\phi}(\cdot\mid \mathbf{x}_{\neg j})} \left[\log \sum_{n=1}^{N} \left( \frac{p_{\theta_{i}} (\mathbf{x}_{i}\mid \mathbf{z}_{\pi}^{n}) \, p(\mathbf{z}_{\pi}^{n})}{q_{\phi}(\mathbf{z}_{\pi}^{n}\mid \mathbf{x}_{\neg i})}  \right)\right].
\label{Eq:conditional-likelihood}
\end{equation}

\section{Additional Cross-Modality Samples}

\begin{figure*}[t]
    \centering
    \begin{subfigure}[b]{0.31\textwidth}
        \centering
        \includegraphics[height=4.3cm]{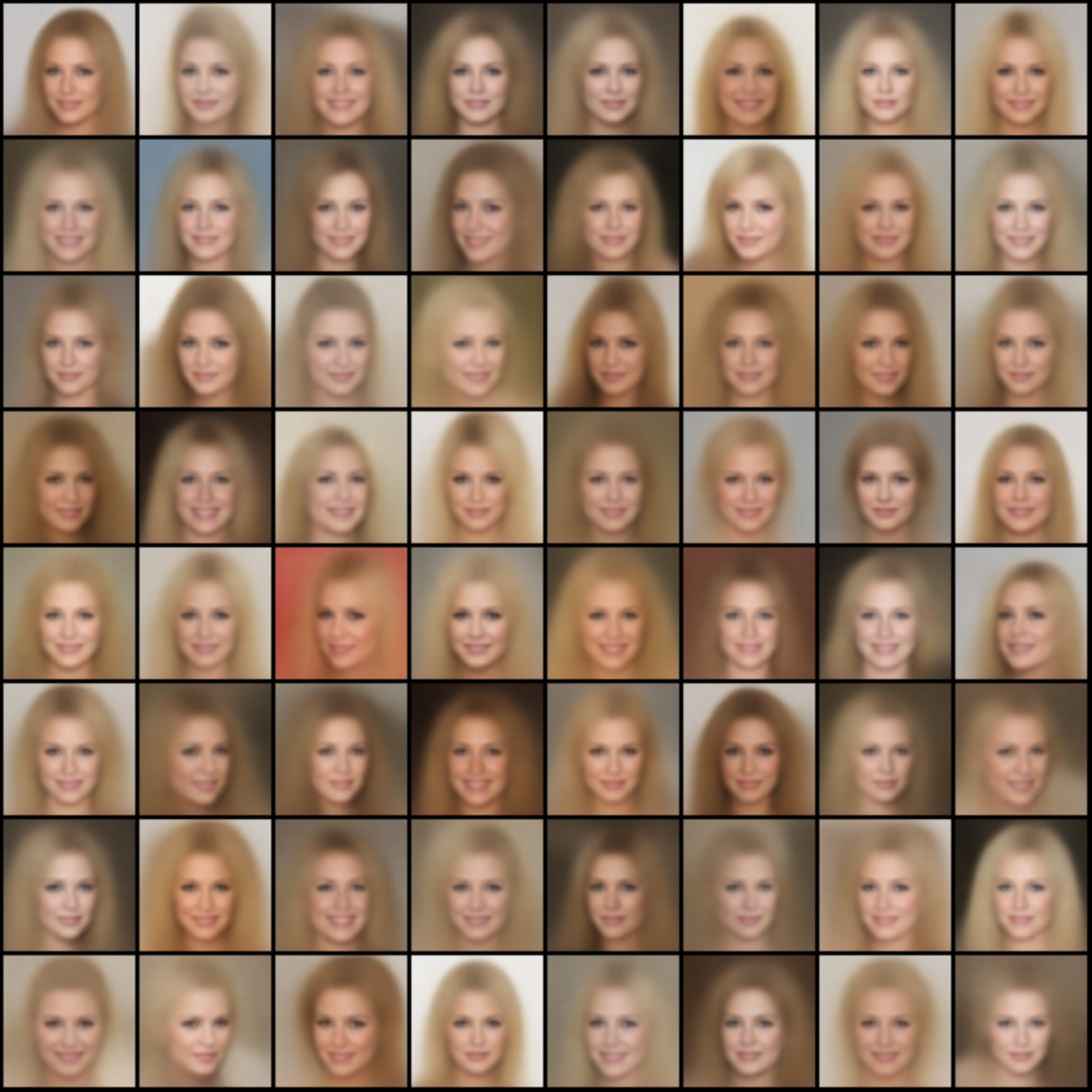}
        \caption{\textbf{MUSE}}
        \label{fig:cross:muse:lady}
    \end{subfigure}%
    \hfill 
    \begin{subfigure}[b]{0.31\textwidth}
        \centering
        \includegraphics[height=4.3cm]{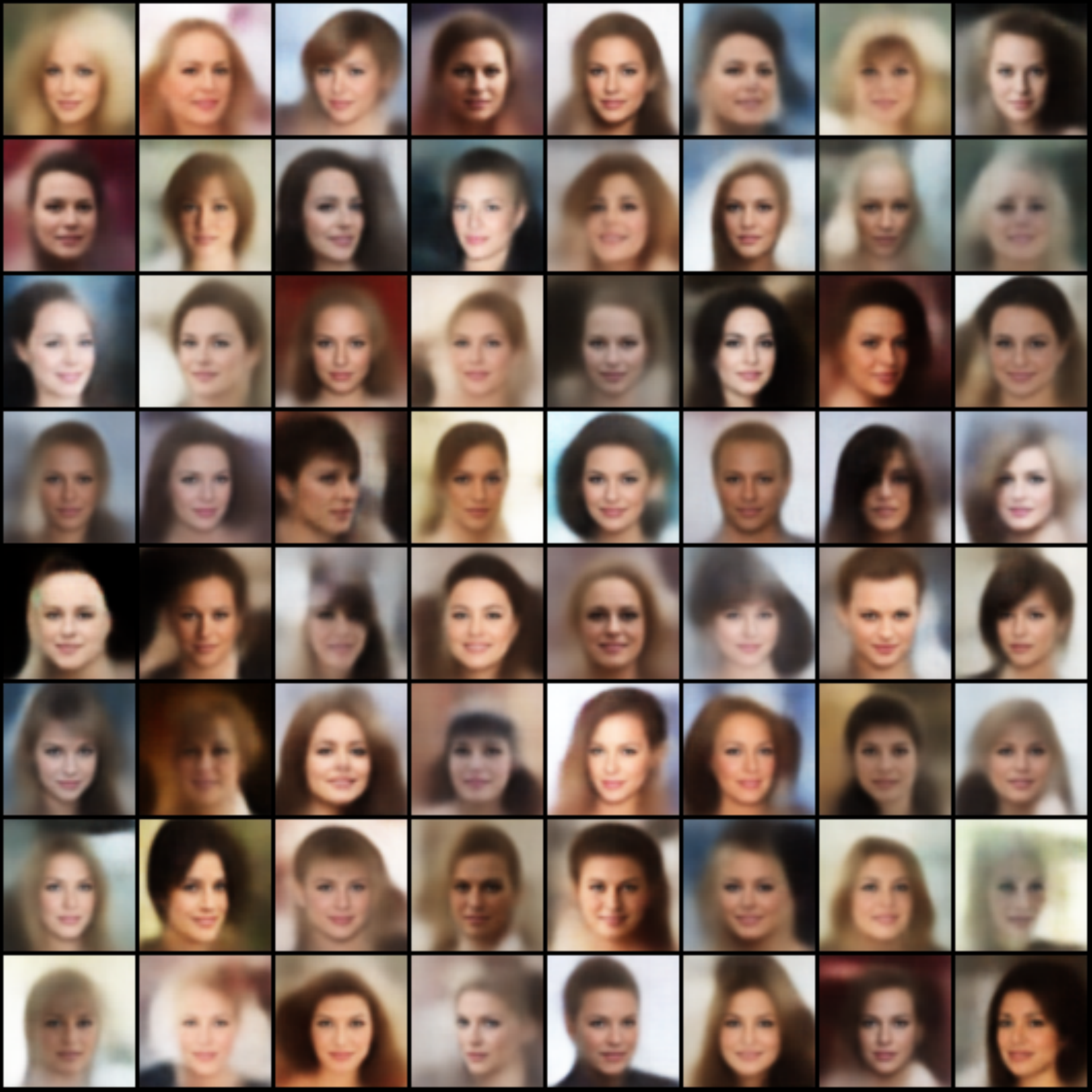}
        \caption{MVAE}
        \label{fig:cross:mvae:lady}
    \end{subfigure}
    \hfill 
    \begin{subfigure}[b]{0.31\textwidth}
        \centering
        \includegraphics[height=4.3cm]{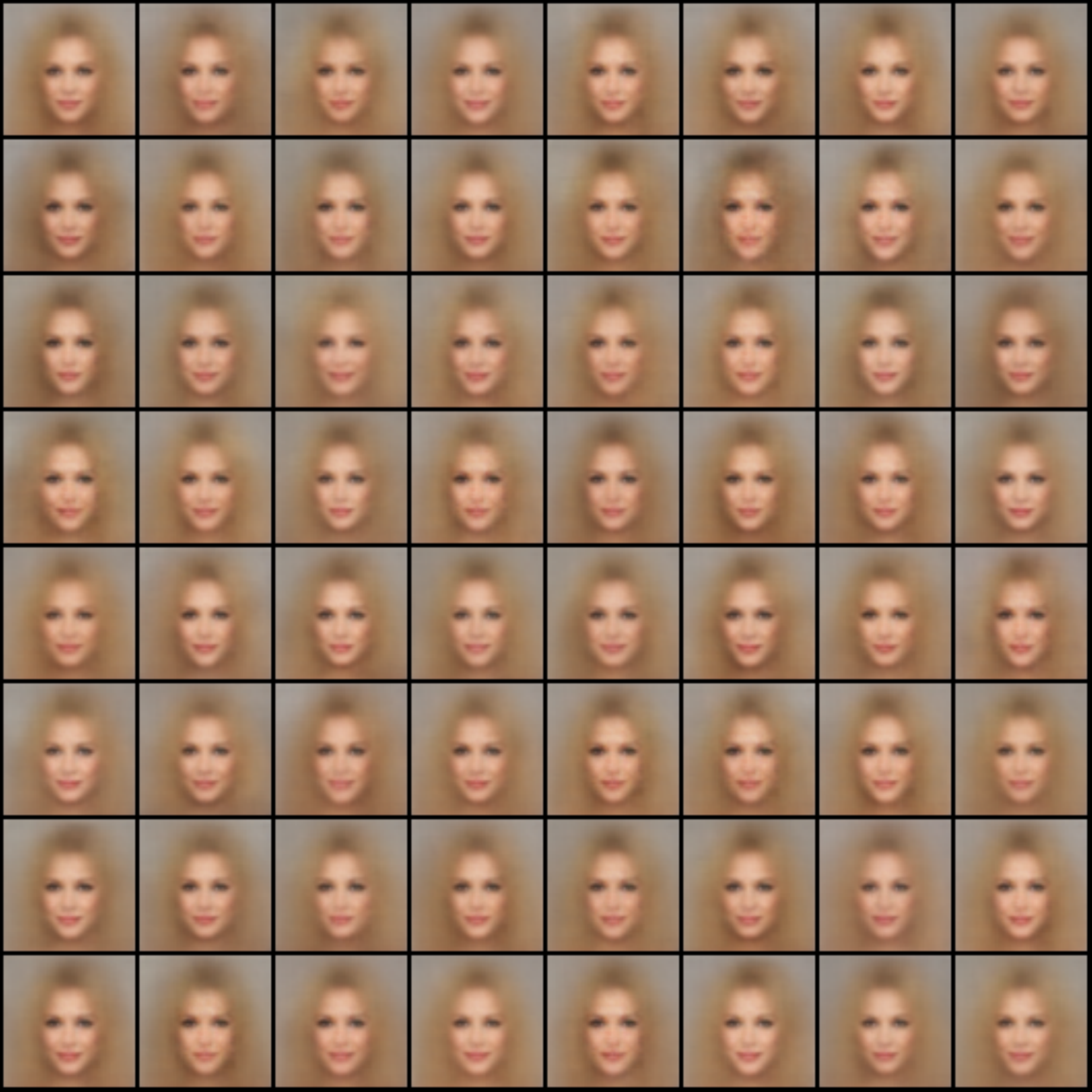}
        \caption{MMVAE}
        \label{fig:cross:mmvae:lady}
    \end{subfigure}
    
    \begin{subfigure}[b]{0.31\textwidth}
        \centering
        \includegraphics[height=4.3cm]{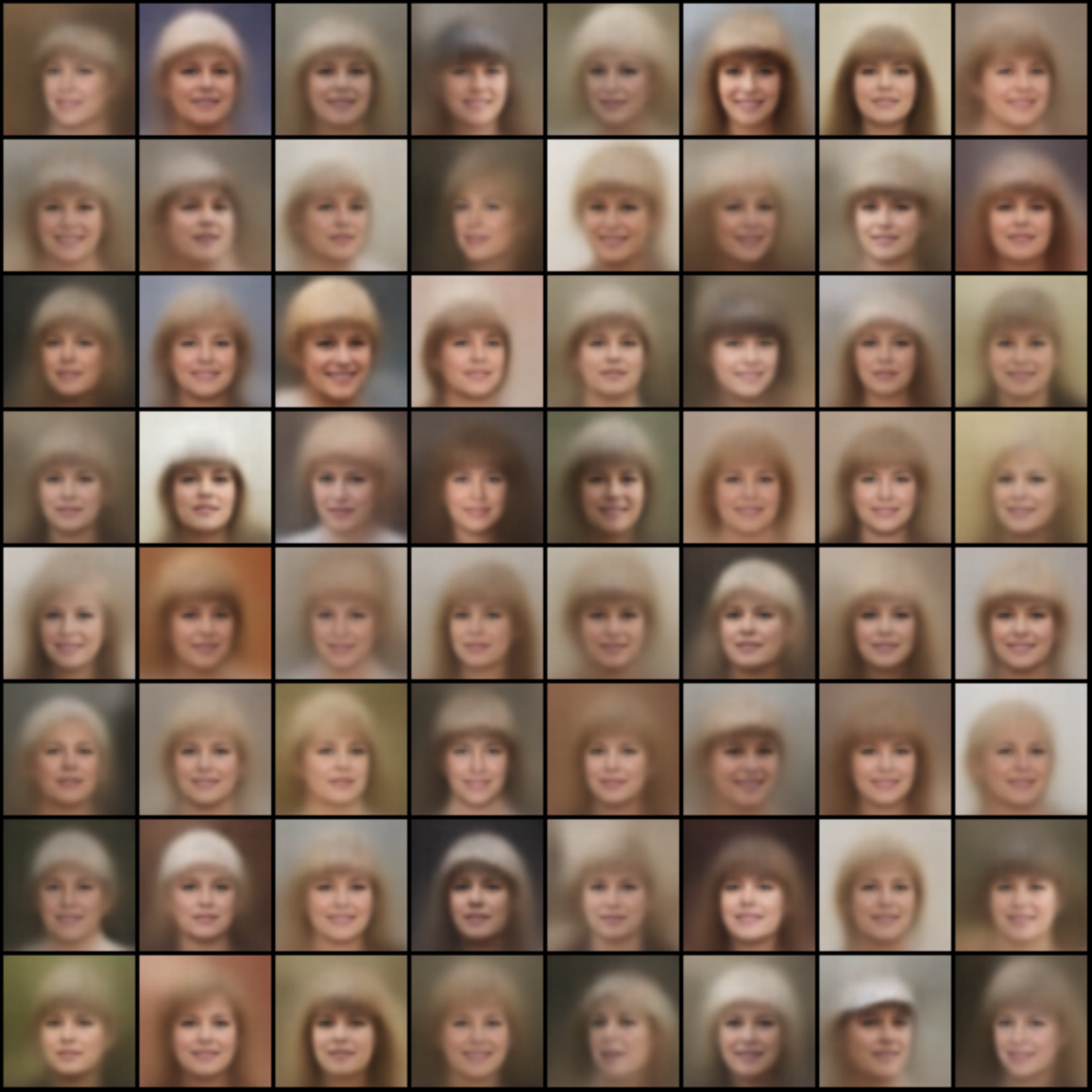}
        \caption{\textbf{MUSE}}
        \label{fig:cross:muse:granny}
    \end{subfigure}%
    \hfill 
    \begin{subfigure}[b]{0.31\textwidth}
        \centering
        \includegraphics[height=4.3cm]{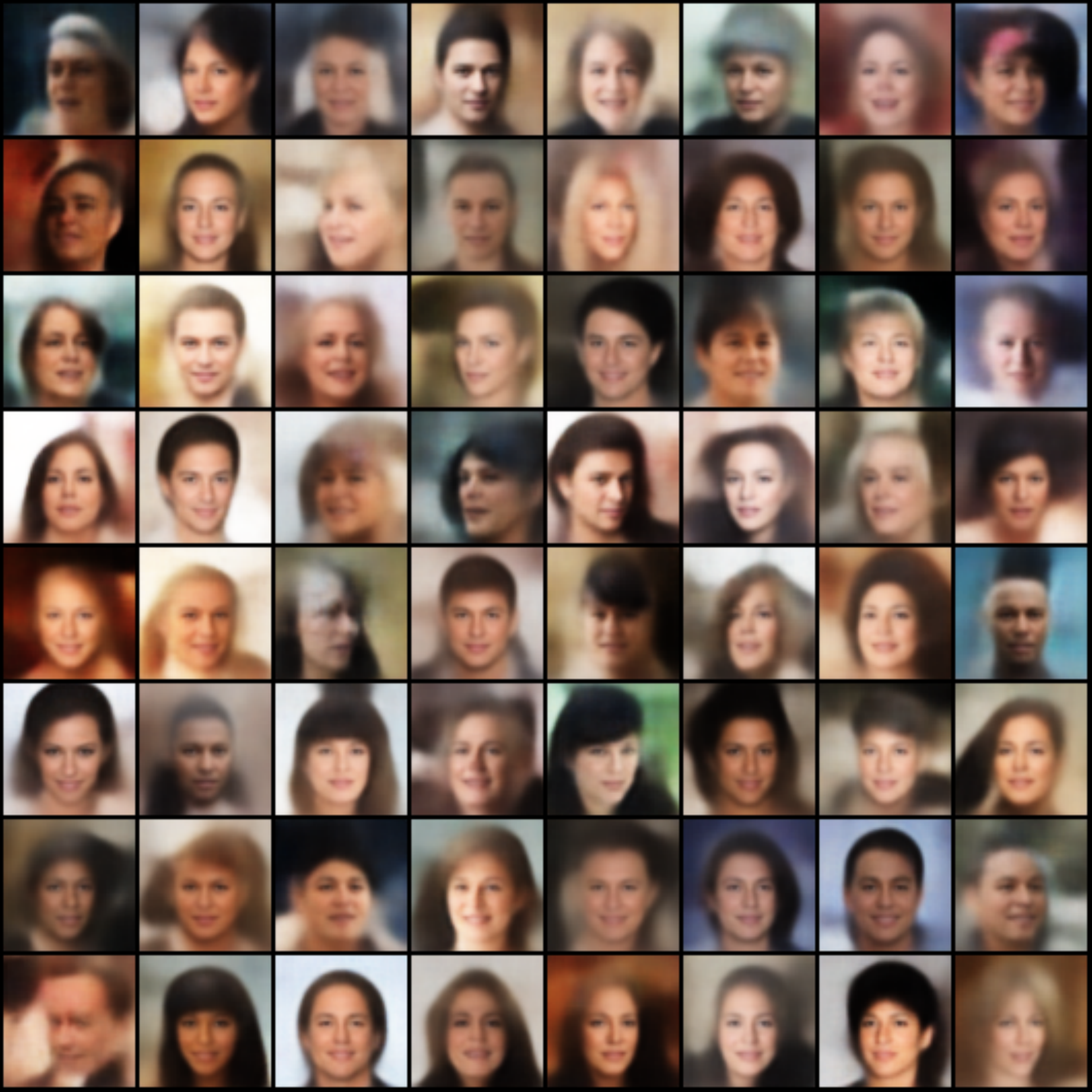}
        \caption{MVAE}
        \label{fig:cross:mvae:granny}
    \end{subfigure}
    \hfill 
    \begin{subfigure}[b]{0.31\textwidth}
        \centering
        \includegraphics[height=4.3cm]{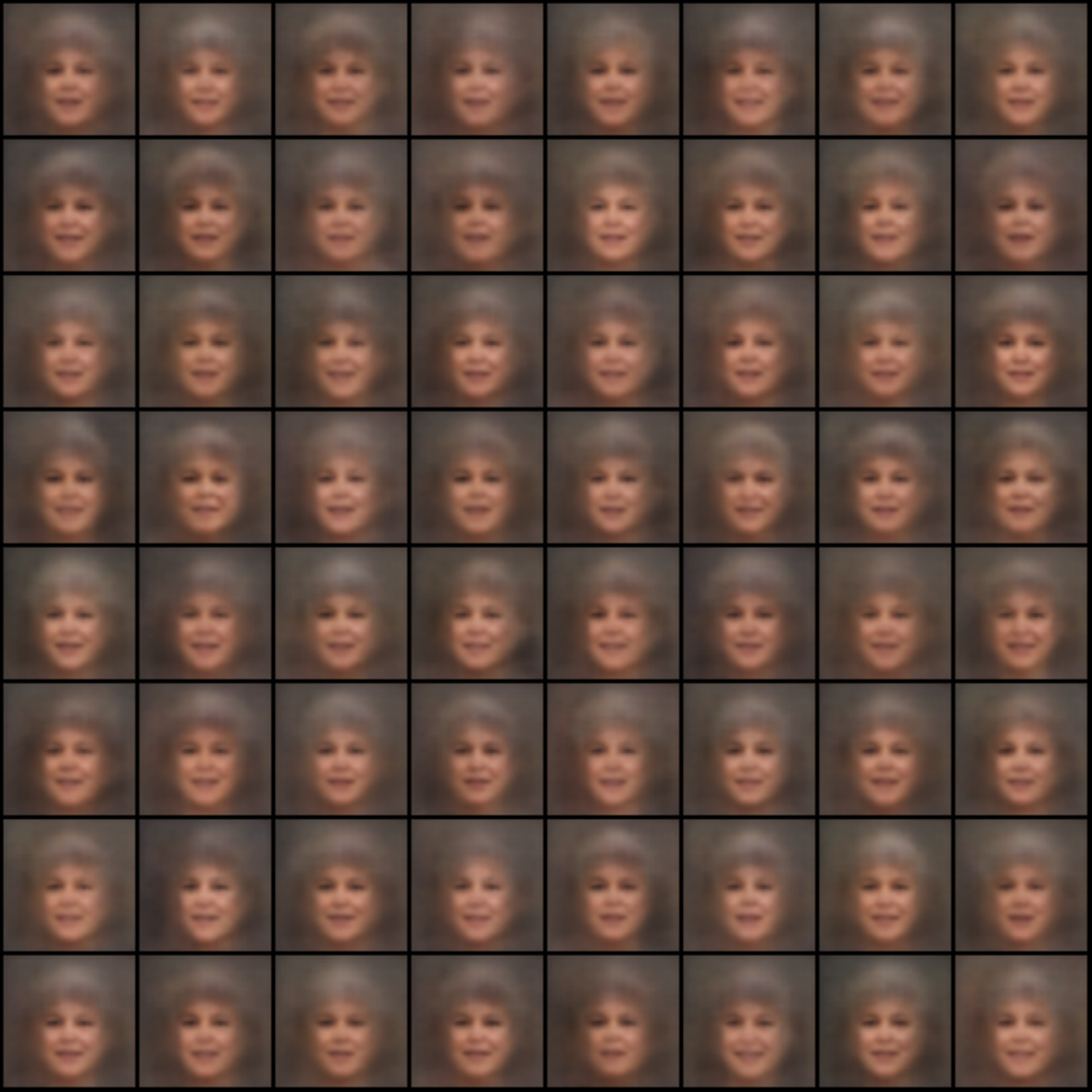}
        \caption{MMVAE}
        \label{fig:cross:mmvae:granny}
    \end{subfigure}
    \caption{Cross-modal image generation from attribute information in the CelebA dataset considering (top row) $\mathbf{x}_2 = \{\text{Female}; \text{Blond Hair}; \text{Heavy Makeup}; \text{Smiling}; \text{Wavy Hair}; \text{Young}\}$ and (bottom row) $\mathbf{x}_2 = \{\text{Female}; \text{Big Nose}; \text{Grey Hair}; \text{Wearing Hat}; \text{Wavy Hair}; \text{Mouth Slightly Open}\}$ (Best viewed with zoom).}
    \label{fig:celeb:samples}
\end{figure*}

We present additional image samples generated from attribute information in the CelebA dataset. The samples, shown in Fig.~\ref{fig:celeb:samples}, attest that only MUSE allows the generation of high-quality, coherent complex information (e.g. images) from low-dimensional information (e.g. labels). 

\section{Description of Multimodal Datasets}

In this section we present the literature-standard multimodal datasets employed in the evaluation of Section~\ref{Section:eval:generative}.

\begin{figure*}[t]
    \centering
    \begin{subfigure}[b]{0.31\textwidth}
        \centering
        \includegraphics[height=4.3cm]{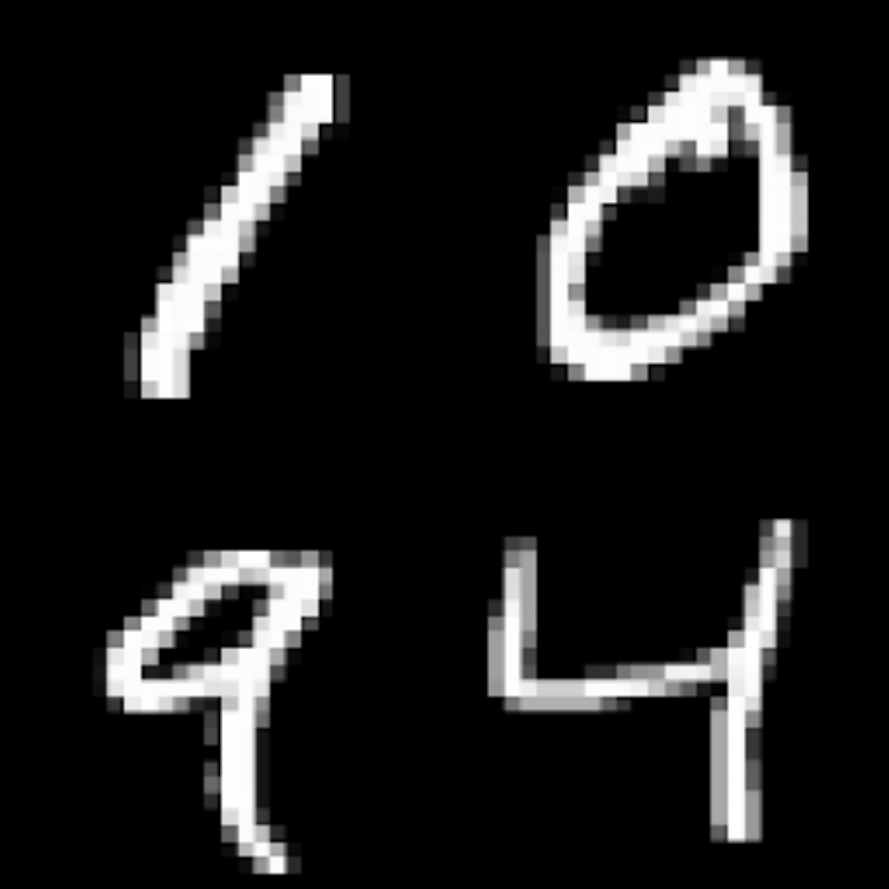}
        \caption{MNIST}
        \label{fig:dataset:mnist}
    \end{subfigure}%
    \hfill 
    \begin{subfigure}[b]{0.31\textwidth}
        \centering
        \includegraphics[height=4.3cm]{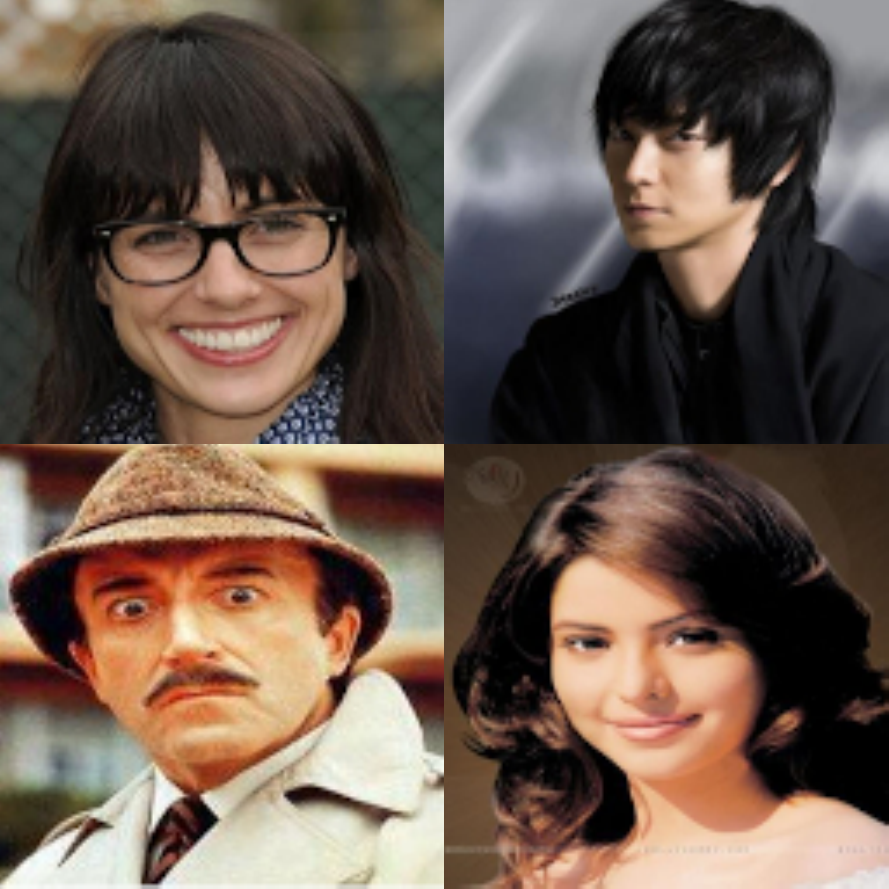}
        \caption{CelebA}
        \label{fig:dataset:celeba}
    \end{subfigure}
    \hfill 
    \begin{subfigure}[b]{0.31\textwidth}
        \centering
        \includegraphics[height=4.3cm]{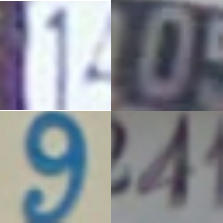}
        \caption{SVHN}
        \label{fig:dataset:svhn}
    \end{subfigure}
    \caption{Image samples from the literature-standard multimodal datasets employed in the evaluation of MUSE of Section 4.2.}
    \label{fig:dataset:samples}
\end{figure*}

\begin{itemize}
    \item The \emph{MNIST} dataset~\cite{lecun1998gradient} is a two-modality scenario ($M=2$):
    \begin{itemize}
        \item $\mathbf{x}_1 \in \mathbb{R}^{1+28+28}$ -  Grayscale image of a handwritten digit (Fig.~\ref{fig:dataset:mnist});
        \item $\mathbf{x}_2 \in \mathbb{R}^{10}$ - Associated digit label.
    \end{itemize}
    We use a total representation space of 64 dimensions for all models. For MUSE, we set the image-specific latent space $\mathbf{z}_1 \in \mathbb{R}^{50}$, the label-specific latent space $\mathbf{z}_2 \in \mathbb{R}^{4}$ and the multimodal latent space $\mathbf{z}_{\pi} \in \mathbb{R}^{10}$;
    \item The \emph{CelebA} dataset~\cite{liu2015faceattributes} is a two-modality scenario ($M=2$):
    \begin{itemize}
        \item $\mathbf{x}_1 \in \mathbb{R}^{3+64+64}$ - RGB image of a human face (Fig.~\ref{fig:dataset:celeba});
        \item $\mathbf{x}_2 \in \mathbb{R}^{40}$ - Associated semantic attribute information.
    \end{itemize}
    We use a total representation space of 100 dimensions for all models. For MUSE, we set the image-specific $\mathbf{z}_1 \in \mathbb{R}^{60}$, the attribute-specific latent space $\mathbf{z}_2 \in \mathbb{R}^{20}$ and the multimodal latent space $\mathbf{z}_{\pi} \in \mathbb{R}^{20}$.
    \item The \emph{MNIST-SVHN} scenario considers three different modalities ($M=3$):
    \begin{itemize}
        \item $\mathbf{x}_1 \in \mathbb{R}^{1+28+28}$ - Grayscale image of a handwritten digit from the MNIST dataset~\cite{lecun1998gradient} (Fig.~\ref{fig:dataset:mnist});
        \item $\mathbf{x}_2 \in \mathbb{R}^{3+ 32+32}$ - Corresponding RGB image of a house numbers from the SVHN dataset~\cite{netzer2011reading} (Fig.~\ref{fig:dataset:svhn});
        \item $\mathbf{x}_3 \in \mathbb{R}^{10}$ - Associated digit label.
    \end{itemize}
    We define a total representation space of 100 dimensions for all models. For MUSE, we set the ``MNIST''-specific latent space $\mathbf{z}_1 \in \mathbb{R}^{40}$, a ``SVHN''-specific latent space $\mathbf{z}_2 \in \mathbb{R}^{40}$, a label-specific latent space $\mathbf{z}_3 \in \mathbb{R}^{4}$, and a multimodal latent space $\mathbf{z}_{\pi} \in \mathbb{R}^{16}$.
\end{itemize}

\section{Training Hyperparameters}
We describe the training hyperparameters employed for the evaluation of MUSE present in Section~\ref{Section:Evaluation}. We recover the total loss of MUSE (Eq.~\ref{Eq:MUSE-loss}), 

\begin{align}
    \ell(\mathbf{x}_{1:M}) = &\sum_{m=1}^{M}  \left(\EX_{q_{\phi}(\mathbf{z}_m | \mathbf{x}_m)} \left[ \lambda_m \log p_{\theta}(\mathbf{x}_m | \mathbf{z}_m) \right] - \alpha_m \KL \infdiv{q_{\phi}(\mathbf{z}_m|\mathbf{x}_m)}{p(\mathbf{z}_m)}\right) \nonumber\\
   &\EX_{q_{\phi}(\mathbf{c}_{1:M}| \mathbf{x}_{1:M})} \left( \beta \KL \infdiv{q_{\phi}(\mathbf{z}_{\pi} \mid \mathbf{c}_{1:M})}{p(\mathbf{z}_{\pi})}
   - \sum_{m=1}^{M} \EX_{q_{\phi}(\mathbf{z}_{\pi} \mid \mathbf{c}_{1:M})} \left[\gamma_m\log p_{\theta} (\mathbf{c}_{m}\mid \mathbf{z}_{\pi})\right]\right),\nonumber\\
   &+ \frac{\delta}{|D|}\sum_{d \in D}\KL^{\star} \infdiv{q_{\phi}(\mathbf{z}_{\pi} \mid \mathbf{c}_{1:M})}{q_{\phi}(\mathbf{z}_{d} \mid \mathbf{c}_{d})},
\end{align}
where the modality-specific hyperparameters $\lambda_m$, $\alpha_m$ and $\gamma_m$ control the modality data reconstruction, modality-specific distribution regularization and modality representation reconstruction objectives, respectively. The hyperparameter $\beta$ controls the regularization of the multimodal latent distribution. We present the training hyperparameters generative employed in the generative evaluation (Section~\ref{Section:eval:generative}) of MUSE in Table~\ref{Table:training_hyperparameters}. For the training hyperparameters employed in the reinforcement learning comparative study (Section~\ref{Section:eval:rl}) please refer to~\citet{silva2020playing}.

\begin{table}[b]
\centering
\caption{Training hyperparameters employed in the evaluation of Section~\ref{Section:eval:generative}.}
\begin{subtable}{0.2\textwidth}
\centering
\caption{MNIST}
\begin{adjustbox}{width=0.9\columnwidth,center}
\begin{tabular}{@{}lc@{}}
\toprule
Parameter & Value \\ \midrule
Training Epochs & 200 \\
Learning Rate & $10^{-3}$ \\
Batch-size & 64 \\
Optimizer & Adam \\
$\lambda_1$ & 1.0 \\
$\lambda_2$ & 50.0 \\
$\alpha_1$ = $\alpha_2$ & 1.0 \\
$\gamma_1$ = $\gamma_2$ & 10.0 \\
$\beta $ & 1.0 \\ 
$\delta$ & 1.0 \\\bottomrule
\end{tabular}
\end{adjustbox}
\end{subtable}
\hfill
\begin{subtable}{0.2\textwidth}
\centering
\caption{CelebA}
\begin{adjustbox}{width=0.9\columnwidth,center}
\begin{tabular}{@{}lc@{}}
\toprule
Parameter & Value \\ \midrule
Training Epochs & 100 \\
Learning Rate & $10^{-4}$ \\
Batch-size & 64 \\
Optimizer & Adam \\
$\lambda_1$ & 1.0 \\
$\lambda_2$ & 50.0 \\
$\alpha_1$ = $\alpha_2$ & 1.0 \\
$\gamma_1$ = $\gamma_2$ & 10.0 \\
$\beta $ & 1.0 \\ 
$\delta$ & 1.0 \\\bottomrule
\end{tabular}
\end{adjustbox}
\end{subtable}
\hfill
\begin{subtable}{0.2\textwidth}
\centering
\caption{MNIST-SVHN}
\begin{adjustbox}{width=0.9\columnwidth,center}
\begin{tabular}{@{}lc@{}}
\toprule
Parameter & Value \\ \midrule
Training Epochs & 100 \\
Learning Rate & $10^{-3}$ \\
Batch-size & 64 \\
Optimizer & Adam \\
$\lambda_1$ = $\lambda_2$ & 1.0 \\
$\lambda_3$ & 50.0 \\
$\alpha_1$ = $\alpha_2$ = $\alpha_3$ & 1.0 \\
$\gamma_1$ = $\gamma_2$ = $\gamma_3$ & 10.0 \\
$\beta $ & 1.0 \\ 
$\delta$ & 1.0 \\\bottomrule
\end{tabular}
\end{adjustbox}
\end{subtable}
\label{Table:training_hyperparameters}
\end{table}

\section{Model Network Architectures}

We now describe the network architectures employed in the generative evaluation of MUSE, presented in Section~\ref{Section:eval:generative}: in Table~\ref{Table:bottom_network}, we present the modality-specific networks and in Table~\ref{Table:top_network} we present the top-level networks, specific for each evaluation. For the network architectures employed in the comparative study of Section~\ref{Section:eval:rl}, please refer to~\citet{silva2020playing}.

\begin{table}[b]
\centering
\caption{Bottom-level, modality-specific network architectures employed in Section 4.2 (best viewed with zoom).}
\begin{subtable}{\textwidth}
\centering
\caption{MNIST - $\mathbf{x}_1$ (Image)}
\begin{adjustbox}{width=0.8\columnwidth,center}
\begin{tabular}{@{}ll@{}}
\toprule
\textbf{Encoder} & \textbf{Decoder} \\ \midrule
Input $\mathbb{R}^{1+28+28}$ & Input  $\mathbb{R}^{D}$ \\
Convolutional, 4x4 kernel, 2 stride, 1 padding + Swish & FC, 512 + Swish \\
Convolutional, 4x4 kernel, 2 stride, 1 padding + Swish & FC, 6272 + Swish \\
FC, 512 + Swish & Transposed Convolutional, 4x4 kernel, 2 stride, 1 padding + Swish \\
FC, $D$, FC, $D$ & Transposed Convolutional, 4x4 kernel, 2 stride, 1 padding + Sigmoid \\ \bottomrule
\end{tabular}
\end{adjustbox}
\end{subtable}

\vspace{2ex}

\begin{subtable}{\textwidth}
\centering
\caption{MNIST - $\mathbf{x}_2$ (Label)}
\begin{adjustbox}{width=0.25\columnwidth,center}
\begin{tabular}{@{}ll@{}}
\toprule
\textbf{Encoder} & \textbf{Decoder} \\ \midrule
Input $\mathbb{R}^{10}$ & Input  $\mathbb{R}^{D}$ \\
FC, 64 + ReLU & FC, 64 + ReLU \\
FC, 64 + ReLU & FC, 64 + ReLU \\
FC, $D$, FC, $D$  & FC, 64 + ReLU \\
- &  FC, 10 + Log Softmax  \\ \bottomrule
\end{tabular}
\end{adjustbox}
\end{subtable}

\vspace{2ex}

\begin{subtable}{\textwidth}
\centering
\caption{CelebA - $\mathbf{x}_1$ (Image)}
\begin{adjustbox}{width=\columnwidth,center}
\begin{tabular}{@{}ll@{}}
\toprule
\textbf{Encoder} & \textbf{Decoder} \\ \midrule
Input $\mathbb{R}^{1+64+64}$ & Input  $\mathbb{R}^{D}$ \\
Convolutional, 4x4 kernel, 2 stride, 1 padding + Swish & FC, 6400 + Swish \\
Convolutional, 4x4 kernel, 2 stride, 1 padding + Batchnorm + Swish & Transposed Convolutional, 4x4 kernel, 1 stride, 0 padding + Batchnorm + Swish \\
Convolutional, 4x4 kernel, 2 stride, 1 padding + Batchnorm + Swish & Transposed Convolutional, 4x4 kernel, 2 stride, 1 padding + Batchnorm + Swish \\
Convolutional, 4x4 kernel, 1 stride, 0 padding + Batchnorm + Swish & Transposed Convolutional, 4x4 kernel, 2 stride, 1 padding + Batchnorm + Swish \\
FC, 512 + Swish + Dropout ($p=0.1)$ & Transposed Convolutional, 4x4 kernel, 2 stride, 1 padding + Sigmoid \\
FC, $D$, FC, $D$ & - \\ \bottomrule
\end{tabular}
\end{adjustbox}
\end{subtable}

\vspace{2ex}

\begin{subtable}{\textwidth}
\centering
\caption{CelebA - $\mathbf{x}_2$ (Attribute)}
\begin{adjustbox}{width=0.37\columnwidth,center}
\begin{tabular}{@{}ll@{}}
\toprule
\textbf{Encoder} & \textbf{Decoder} \\ \midrule
Input $\mathbb{R}^{40}$ & Input  $\mathbb{R}^{D}$ \\
FC, 512 + Batchnorm + Swish & FC, 512 + Batchnorm \\
FC, 512 + Batchnorm + Swish & FC, 512 + Batchnorm \\
FC, $D$, FC, $D$  & FC, 512 + Batchnorm \\
- &  FC, 40 + Sigmoid  \\ \bottomrule
\end{tabular}
\end{adjustbox}
\end{subtable}

\vspace{2ex}

\begin{subtable}{\textwidth}
\centering
\caption{SVHN - $\mathbf{x}_2$ (Image)}
\begin{adjustbox}{width=0.85\columnwidth,center}
\begin{tabular}{@{}ll@{}}
\toprule
\textbf{Encoder} & \textbf{Decoder} \\ \midrule
Input $\mathbb{R}^{1+64+64}$ & Input  $\mathbb{R}^{D}$ \\
Convolutional, 4x4 kernel, 2 stride, 1 padding + Swish & Transposed Convolutional, 4x4 kernel, 1 stride, 0 padding + Swish \\
Convolutional, 4x4 kernel, 2 stride, 1 padding + Swish & Transposed Convolutional, 4x4 kernel, 2 stride, 1 padding + Swish \\
Convolutional, 4x4 kernel, 2 stride, 1 padding + Swish & Transposed Convolutional, 4x4 kernel, 2 stride, 1 padding + Swish \\
FC, 1024 + Swish & Transposed Convolutional, 4x4 kernel, 2 stride, 1 padding + Sigmoid \\
FC, 512 + Swish & - \\
FC, $D$, FC, $D$ & - \\ \bottomrule
\end{tabular}
\end{adjustbox}
\end{subtable}

\label{Table:bottom_network}
\end{table}

\begin{table}[t]
\centering
\caption{Top-level, multimodal network architectures, where $D_m = |\mathbf{z}_m|$ and $D_{\pi} = |\mathbf{z}_{\pi}|$, employed in Section 4.2 (best viewed with zoom).}
\begin{subtable}{0.25\textwidth}
\centering
\caption{MNIST}
\begin{adjustbox}{width=\columnwidth,center}
\begin{tabular}{@{}ll@{}}
\toprule
\textbf{Encoder} & \textbf{Decoder} \\ \midrule
Input $\mathbb{R}^{D_m}$ & Input  $\mathbb{R}^{D_{\pi}}$ \\
FC, 128 + ReLU & FC, 128 + ReLU \\
FC, 128 + ReLU & FC, 128 + ReLU \\
FC, $D$$_{\pi}$, FC, $D$$_{\pi}$ & FC, D$_m$ \\ \bottomrule
\end{tabular}
\end{adjustbox}
\end{subtable}
\hfill
\begin{subtable}{0.25\textwidth}
\centering
\caption{CelebA}
\begin{adjustbox}{width=\columnwidth,center}
\begin{tabular}{@{}ll@{}}
\toprule
\textbf{Encoder} & \textbf{Decoder} \\ \midrule
Input $\mathbb{R}^{D_m}$ & Input  $\mathbb{R}^{D_{\pi}}$ \\
FC, 128 + ReLU & FC, 128 + ReLU \\
FC, 128 + ReLU & FC, 128 + ReLU \\
FC, $D$$_{\pi}$, FC, $D$$_{\pi}$ & FC, $D$$_m$ \\ \bottomrule
\end{tabular}
\end{adjustbox}
\end{subtable}
\hfill
\begin{subtable}{0.25\textwidth}
\centering
\caption{MNIST-SVHN}
\begin{adjustbox}{width=\columnwidth,center}
\begin{tabular}{@{}ll@{}}
\toprule
\textbf{Encoder} & \textbf{Decoder} \\ \midrule
Input $\mathbb{R}^{D_m}$ & Input  $\mathbb{R}^{D_{\pi}}$ \\
FC, 512 + ReLU & FC, 512 + ReLU \\
FC, 512 + ReLU & FC, 512 + ReLU \\
FC, 512 + ReLU & FC, 512 + ReLU \\
FC, $D$$_{\pi}$, FC, $D$$_{\pi}$ & FC, $D$$_m$ \\ \bottomrule
\end{tabular}
\end{adjustbox}
\end{subtable}
\label{Table:top_network}
\end{table}

\end{document}